\documentclass{article}

\usepackage[preprint]{neurips_2024}

\usepackage[utf8]{inputenc} 
\usepackage[T1]{fontenc}    
\usepackage{hyperref}       
\usepackage{url}            
\usepackage{booktabs}       
\usepackage{amsfonts}       
\usepackage{nicefrac}       
\usepackage{microtype}      
\usepackage[dvipsnames]{xcolor}         
\usepackage{times}
\usepackage{latexsym}
\usepackage{CJKutf8}
\usepackage{amsmath}
\usepackage{verbatim}
\usepackage{booktabs}
\usepackage{bbm}
\usepackage{amssymb}
\usepackage{pifont}
\usepackage{stfloats}
\usepackage{multicol}
\usepackage{float}
\usepackage{graphicx}
\usepackage{subfigure}
\usepackage{multirow}
\usepackage{array}
\usepackage{enumerate}
\newcommand{\PreserveBackslash}[1]{\let\temp=\\#1\let\\=\temp}
\newcolumntype{C}[1]{>{\PreserveBackslash\centering}p{#1}}
\newcolumntype{R}[1]{>{\PreserveBackslash\raggedleft}p{#1}}
\newcolumntype{L}[1]{>{\PreserveBackslash\raggedright}p{#1}}
\usepackage{diagbox}
\usepackage{colortbl}
\usepackage{enumitem}
\usepackage{bm}
\usepackage{mathtools}
\usepackage{wrapfig}
\usepackage[skins]{tcolorbox}
\tcbuselibrary{breakable} 
\newtcolorbox[auto counter, number within=section, list type=subsubsection, list inside=toc]{sectionbox}[1]{
colback=white, colframe=black, 
colbacktitle=white!80!gray, coltitle=black, 
fonttitle=\bfseries, title={Comment \thetcbcounter}, list entry={Comment \thetcbcounter\quad}, 
breakable, 
before upper={\parindent10pt\noindent},  
left = 1mm, 
    right = 1mm,
    top = 1mm,
    bottom = 1mm,
}

\usepackage{CJKutf8}
\usepackage{tabularx}
\usepackage{amsmath}
\usepackage{ragged2e}
\usepackage{longtable}
\usepackage{lscape}
\newcommand\icon{\raisebox{-3.7pt}{\includegraphics[width=1.1em]{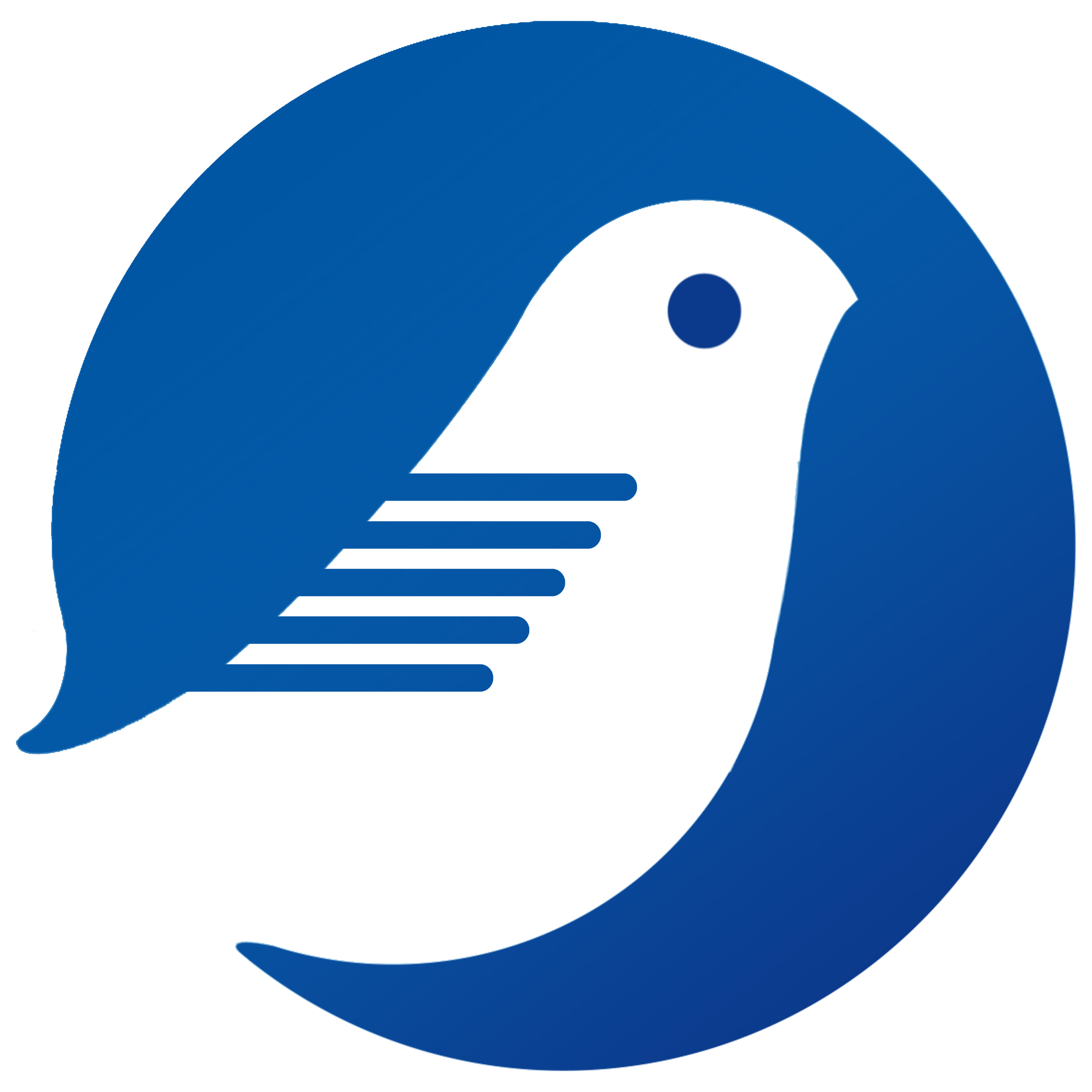}}}

\title{\icon~BayLing 2: A Multilingual Large Language Model with Efficient Language Alignment}

\author{
Shaolei Zhang\textsuperscript{\rm 1,3},\quad
Kehao Zhang\textsuperscript{\rm 1,3},\quad
Qingkai Fang\textsuperscript{\rm 1,3},\quad
Shoutao Guo\textsuperscript{\rm 1,3},\quad
Yan Zhou\textsuperscript{\rm 1,3},\quad\\
\textbf{Xiaodong Liu\textsuperscript{\rm 4},\quad Yang Feng\textsuperscript{\rm 1,2,3}\thanks{Corresponding author: Yang Feng.}} \\
\textsuperscript{\rm 1}{Key Laboratory of Intelligent Information Processing,} \\ Institute of Computing Technology, Chinese Academy of Sciences (ICT/CAS) \\
{ \textsuperscript{\rm 2} {Key Laboratory of AI Safety, Chinese Academy of Sciences}} \\
{ \textsuperscript{\rm 3} {University of Chinese Academy of Sciences, Beijing, China}} \\
\textsuperscript{\rm 4}{Research Center of Distributed Systems,} \\ Institute of Computing Technology, Chinese Academy of Sciences (ICT/CAS) \\
\texttt{\{\href{mailto:zhangshaolei20z@ict.ac.cn}{zhangshaolei20z},\href{mailto:fengyang@ict.ac.cn}{fengyang}\}@ict.ac.cn}}

\begin{document}

\maketitle

\begin{abstract}

Large language models (LLMs), with their powerful generative capabilities and vast knowledge, empower various tasks in everyday life. However, these abilities are primarily concentrated in high-resource languages, leaving low-resource languages with weaker generative capabilities and relatively limited knowledge. Enhancing the multilingual capabilities of LLMs is therefore crucial for serving over 100 linguistic communities worldwide. An intuitive approach to enhance the multilingual capabilities would be to construct instruction data for various languages, but constructing instruction data for over 100 languages is prohibitively costly. 
In this paper, we introduce BayLing 2, which efficiently transfers generative capabilities and knowledge from high-resource languages to low-resource languages through language alignment. To achieve this, we constructed a dataset of 3.2 million instructions, comprising high-resource language instructions (Chinese and English) and cross-lingual instructions for 100+ languages and performed instruction tuning based on the dataset to facilitate the capability transfer between languages.
Using Llama as the foundation model, we developed BayLing-2-7B, BayLing-2-13B, and BayLing-2-8B, and conducted a comprehensive evaluation of BayLing. For multilingual translation across 100+ languages, BayLing shows superior performance compared to open-source models of similar scale. For multilingual knowledge and understanding benchmarks, BayLing achieves significant improvements across over 20 low-resource languages, demonstrating its capability of effective knowledge transfer from high-resource to low-resource languages. Furthermore, results on English benchmarks indicate that BayLing maintains high performance in high-resource languages while enhancing the performance in low-resource languages. Demo\footnote{Demo: \href{http://nlp.ict.ac.cn/bayling/demo}{http://nlp.ict.ac.cn/bayling/demo}\qquad \textsuperscript{\rm3}Homepage: \href{https://aicloud.oneainexus.cn:30013/bayling2/\#/bayling2}{http://nlp.ict.ac.cn/bayling}}, homepage\footnotesize\footnotemark, code\footnote{Code: \href{https://github.com/ictnlp/BayLing}{https://github.com/ictnlp/BayLing}\qquad $\!\!$ \textsuperscript{\rm5}Models: \href{https://huggingface.co/ICTNLP/bayling-2-7b}{BayLing-2-7B}, \href{https://huggingface.co/ICTNLP/bayling-2-13b}{BayLing-2-13B}, \href{https://huggingface.co/ICTNLP/bayling-2-llama-3-8b}{BayLing-2-8B}} and models\footnotesize\footnotemark of BayLing are available.

\end{abstract}

\section{Introduction}

In recent years, the field of natural language processing (NLP) has witnessed a significant surge in the development and utilization of large language models (LLMs) \citep{chatgpt,openai2023gpt4}. Equipped with rich knowledge, strong generative capabilities, and diverse instruction-following abilities, LLMs empower various specific tasks such as translation, summarization, chat and question answering, seamlessly integratd into everyday life.

However, a significant portion of these potent capabilities is primarily concentrated in English, stemming from the fact that high-resource languages, with English as the representative, occupy over 90\% of the pre-training and fine-tuning corpora \citep{chang2023multilinguality,touvron2023llama,xu2024survey}. This results in issues such as lack of knowledge and lower generative capabilities in many other low-resource languages \citep{nguyen2023seallms,alabi2022adapting}. It is imperative to recognize that linguistic diversity is a fundamental aspect of human communication, with over 7000 spoken languages worldwide, and more than 200 of them are writable \citep{chang2023multilinguality}. The accelerating force of globalization underscores the importance of leveraging LLMs to serve diverse linguistic communities.

Enhancing the multilingual capabilities of LLMs is not a trivial task. The intuitive approach is to construct instruction data for various languages to enhance the LLM's ability to follow instructions and generate responses across different languages \citep{zeng2023greenplm}. However, given the extremely limited instruction data available for some low-resource languages and the prohibitive manual efforts required to construct instructions for over 100 languages, this approach becomes impractical. Therefore, exploring more efficient approaches to improving the performance of LLMs across diverse languages remains an area for further investigation.

On these grounds, we developed BayLing 2, a multilingual LLM, which transfers knowledge, generative capability and instruction-following ability from high-resource to low-resource languages through fine-tuning LLMs on cross-lingual tasks. Previously, BayLing 1 successfully explored transferring English knowledge and capabilities to Chinese through cross-lingual alignment \citep{zhang2023bayling}. Building upon BayLing 1, BayLing 2 extends language alignment to multilingual settings, particularly between high-resource and low-resource languages, leading to a multilingual LLM. The fine-tuning corpus of BayLing 2 primarily consists of Chinese and English instructions, supplemented with rich cross-lingual task instructions between Chinese/English and over 100 other languages, facilitating the capability transfer across languages.

Based on foundational models Llama-2-7B-Chat, Llama-2-13B-Chat, and Llama-3-8B-Instruct, we developed BayLing-2-7B, BayLing-2-13B, and BayLing-2-8B through the proposed efficient language alignment. We conducted a comprehensive evaluation of BayLing's performance on both multilingual and general tasks and assessed the quality of language alignment through multilingual translation using the Flores-101 and WMT22 benchmarks. BayLing showed superior translation performance across more than 100 languages, achieving the best results among open-source models of comparable scale. We further evaluated BayLing's multilingual knowledge and generative capabilities using benchmarks including Belebele, Multilingual HellaSwag, XNLI, and Multilingual ARC. The results indicated significant performance improvements across more than 20 low-resource languages, such as Bambara, Luganda, Swahili, and Zulu. This demonstrates effective knowledge and generative capability transfer from high-resource to low-resource languages. Additionally, we evaluated BayLing on various general benchmarks (primarily in English), finding that the language alignment had minimal impacts on BayLing's performance in high-resource languages.

By further analyzing the experimental results, we get the following findings:
\begin{itemize}[,itemsep=0pt,topsep=0pt,leftmargin=12pt]
\setlength{\itemsep}{0pt}
\setlength{\parsep}{0pt}
\setlength{\parskip}{0pt}
    \item By fine-tuning on high-resource language instructions and cross-lingual instructions, LLM can transfer knowledge and generative capabilities from high-resource languages to low-resource languages, thereby facilitating multilingual interaction.
    \item Cross-lingual instructions, such as interactive translation and multilingual translation, can efficiently enhance the language alignment within LLM, thereby improving translation performance.
    \item Fine-tuning LLM solely on high-resource language instructions will involve inter-language conflicts and significantly impair the multilingual capabilities of LLM, especially on the low-resource languages. Beside high-resource language instructions, introducing cross-lingual instructions can effectively solve this issue.
\end{itemize}

\begin{figure}[t]
    \centering
    \includegraphics[width=\textwidth]{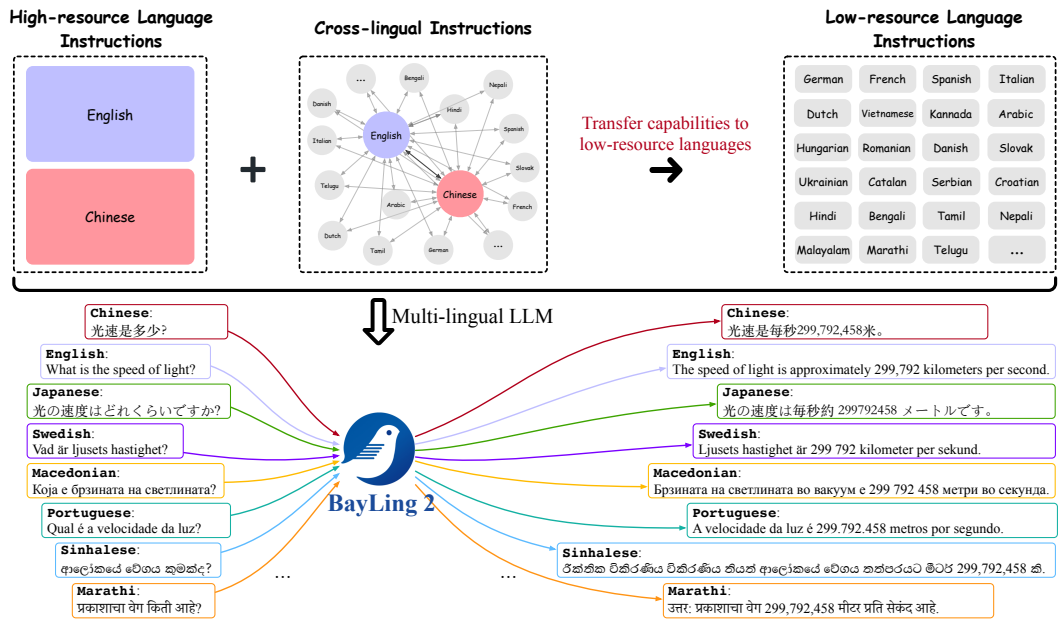}
    \caption{Overview of BayLing 2. BayLing 2 is a multilingual LLM with efficient language alignment. BayLing 2 designates Chinese and English, two high-resource languages, as pivot languages and applies cross-lingual tasks to align 100+ languages to these pivot languages, which facilitates the capabilities transfer from high-resource languages to low-resource languages. During inference, BayLing 2 is capable of high-quality interaction across multiple languages.}
    \label{fig:model}
\end{figure}

\section{Related Work}

Multilingual LLMs, with their capability to handle and produce content in multiple languages simultaneously, hold promise for serving diverse linguistic communities. Foundational models, such as Llama \citep{touvron2023llama}, GPT-3 \citep{brown2020language}, PaLM \citep{chowdhery2022palm}, OPT \citep{zhang2022opt} and GLM \citep{du-etal-2022-glm}, are pretrained on corpora sourced from the web and books, which often encompass multiple languages. However, the distribution of languages in these corpora is notably imbalanced. Specifically, a few high-resource languages dominate a significant portion of the corpus, while a vast number of low-resource languages occupy only a small fraction \citep{touvron2023llama}. This leads to performance variations across different languages \citep{ojo2023good,nguyen2023seallms}. Moreover, subsequent supervised fine-tuning on English-centric instruction data exacerbates the issue of language imbalance \citep{lai2023okapi}, rendering LLMs lower interactive capability with low-resource languages.

Current approaches mainly fall into two categories: continual pretraining and supervised fine-tuning. With continual pretraining, some works focus on continuously pretraining foundational models using multilingual corpora to enhance their multilingual capabilities \citep{nguyen2023culturax,lai2023okapi,ke2023continual,gupta2023continual}. These approaches effectively supplements LLMs with multilingual knowledge and generation abilities. However, continual pretraining often relies on large amounts of multilingual data, and thereby the costs associated with data collection and training are significant \citep{nguyen2023culturax,liu2024translation}. Moreover, there is a risk of catastrophic forgetting with continual pretraining, which may compromise the performance of the foundational model on high-resource languages \citep{li2024quantifying}. Additionally, since the pretraining corpora of foundational models are often close-sourced, it is challenging to maintain the same distribution between the continual pretraining data and the pretraining data, which may lead to conflicting knowledge and potential hallucinations.

For supervised fine-tuning, existing methods attempt to manually annotate multilingual instructions to activate LLMs' ability for multilingual interaction \citep{eisenschlos2020multifit,alabi2022adapting,lai2023okapi,wang2024ultralink,shaham2024multilingual}. This approach often relies on manually annotation and overlooks leveraging the capabilities of foundational models in high-resource languages as well as the generalization ability of LLMs. To address this, BayLing 2 attempts to enhance the multilingual capabilities of LLMs in a more efficient manner. The instruction dataset of BayLing 2 comprises instructions in both high-resource languages and cross-lingual instructions. The instructions in high-resource languages are designed to activate LLMs' instruction-following capability, while cross-lingual instructions aim to facilitate multilingual alignment of LLMs, thereby transferring knowledge, instruction-following, and generation abilities from high-resource languages to low-resource languages.

\section{BayLing 2}

We introduce BayLing 2, a LLM equipped with enhanced multilingual capabilities through efficient language alignment. Building upon open-source foundational models, BayLing 2 endeavors to explore an efficient and cost-effective approach to enhance the multilingual capabilities, thereby addressing the demands for multilingual interaction.

\subsection{Multilingual Alignments with Cross-lingual Tasks}
\label{sec:Instruction}

During the pre-training stage, the distribution of languages in the corpus is highly imbalanced. For instance, English, being a high-resource language, accounts for over 90\% of the corpus, while low-resource languages such as Sinhalese, Marathi, and Macedonian collectively comprise less than 1\% of the corpus. Naturally, foundational models trained on such language-imbalanced corpus exhibit superior performance on English compared to low-resource languages. Previous studies have often noted that due to the generalization capability, LLMs also demonstrate a certain advantage on those languages within the same language family as English. Naturally, aligning low-resource languages with high-resource languages already mastered by LLMs allows us to transfer the knowledge and generation capabilities of LLMs from high-resource languages to other languages efficiently, thereby enhancing the multilingual capabilities of LLMs.

We employ cross-lingual tasks to align low-resource languages with high-resource languages, thereby achieving multilingual alignment. Fortunately, translation tasks naturally serve as well-defined cross-lingual tasks, demanding outputs that maintain consistent meanings with inputs while differing in language. More importantly, translation tasks boast abundant high-quality parallel corpus across diverse domains, thus laying the groundwork for the efficient achievement of language alignment.

\begin{figure}[t]
    \centering
    \includegraphics[width=0.99\textwidth]{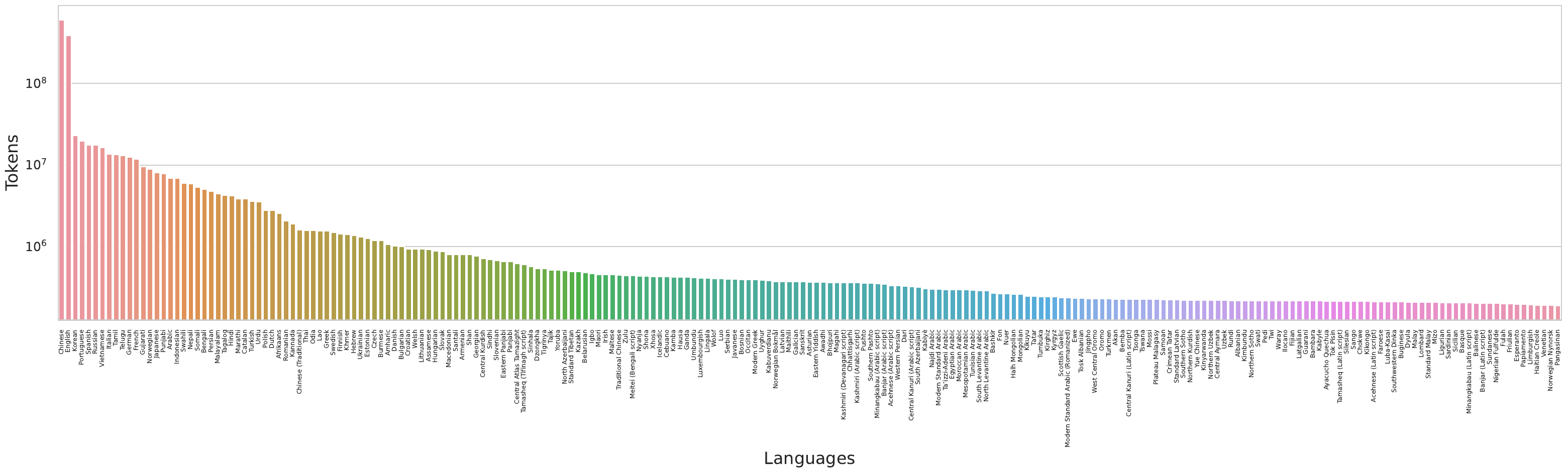}
    \caption{Language distribution of instruction dataset.}
    \label{fig:lang}
\end{figure}

\begin{figure}[t]
    \centering
    \begin{minipage}{0.3\textwidth}
        \centering
        \includegraphics[width=\textwidth]{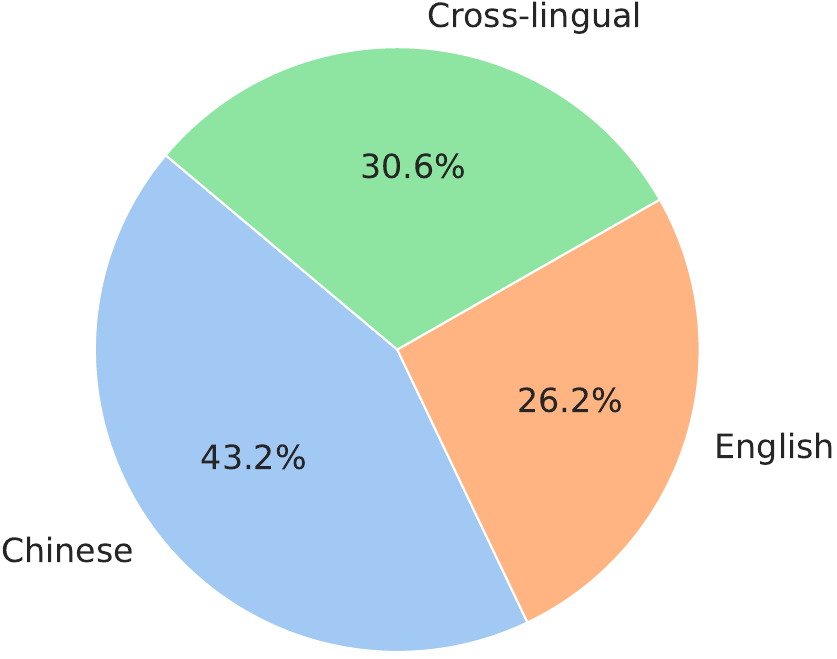} 
        \caption{Distribution of instruction categories, including Chinese, English and cross-lingual instructions.}
        \label{fig:type}
    \end{minipage}\quad \quad
    \begin{minipage}{0.63\textwidth}
        \centering
        \includegraphics[width=\textwidth]{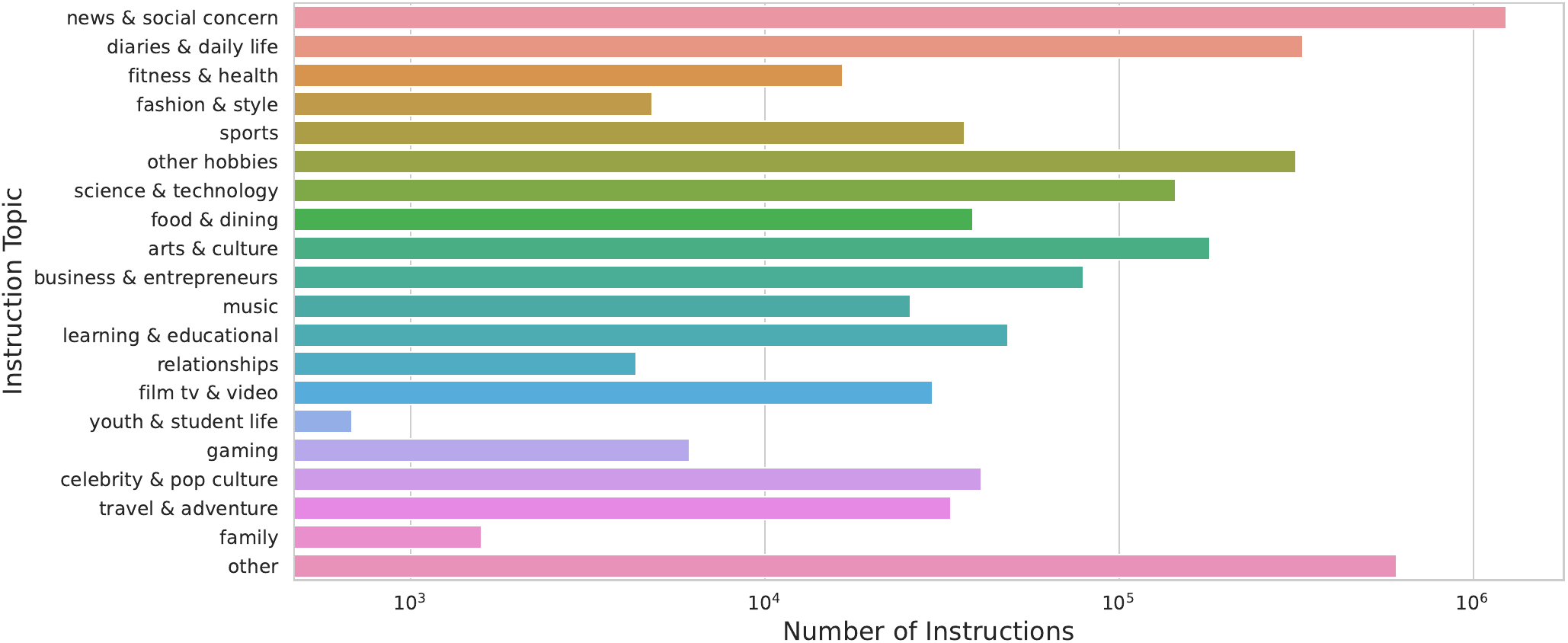} 
        \caption{Distribution of the tokens number involved in each instruction.}
        \label{fig:topic}
    \end{minipage}
\end{figure}

Specifically, we designate Chinese and English, two high-resource languages, as the pivot language, and align over 100 other languages to Chinese and English using translation instructions. Following this idea, we construct the instruction dataset for BayLing 2, comprising Chinese instructions, English instructions and cross-lingual instructions, as shown in Figure~\ref{fig:model}. The instruction dataset contains a total of 3.2 million instructions (1471 million tokens), with the distribution of Chinese, English and cross-lingual instructions illustrated in Figure~\ref{fig:type}. Notably, the cross-lingual instructions in BayLing 2 involve interactive translation, constrained translation, document-level translation and single-sentence translation tasks across over 100 languages. The language distribution of the instruction dataset is shown in Figure~\ref{fig:lang}. The distribution of topics covered in the proposed instruction dataset is illustrated in Figure~\ref{fig:topic}, where instructions primarily sourced from news corpora ensure data quality and security. Overall, BayLing 2 is fine-tuned on 3.2 million instructions covering 100+ languages, achieving multilingual alignment and thereby transferring knowledge and generation capabilities from high-resource languages to low-resource languages.

\subsection{Training}

\setlength{\columnsep}{11pt}
\begin{wrapfigure}{r}{0.47\textwidth}
\begin{center}
\advance\leftskip+1mm
 \vspace{-0.05in} 
 \includegraphics[width=2.4in]{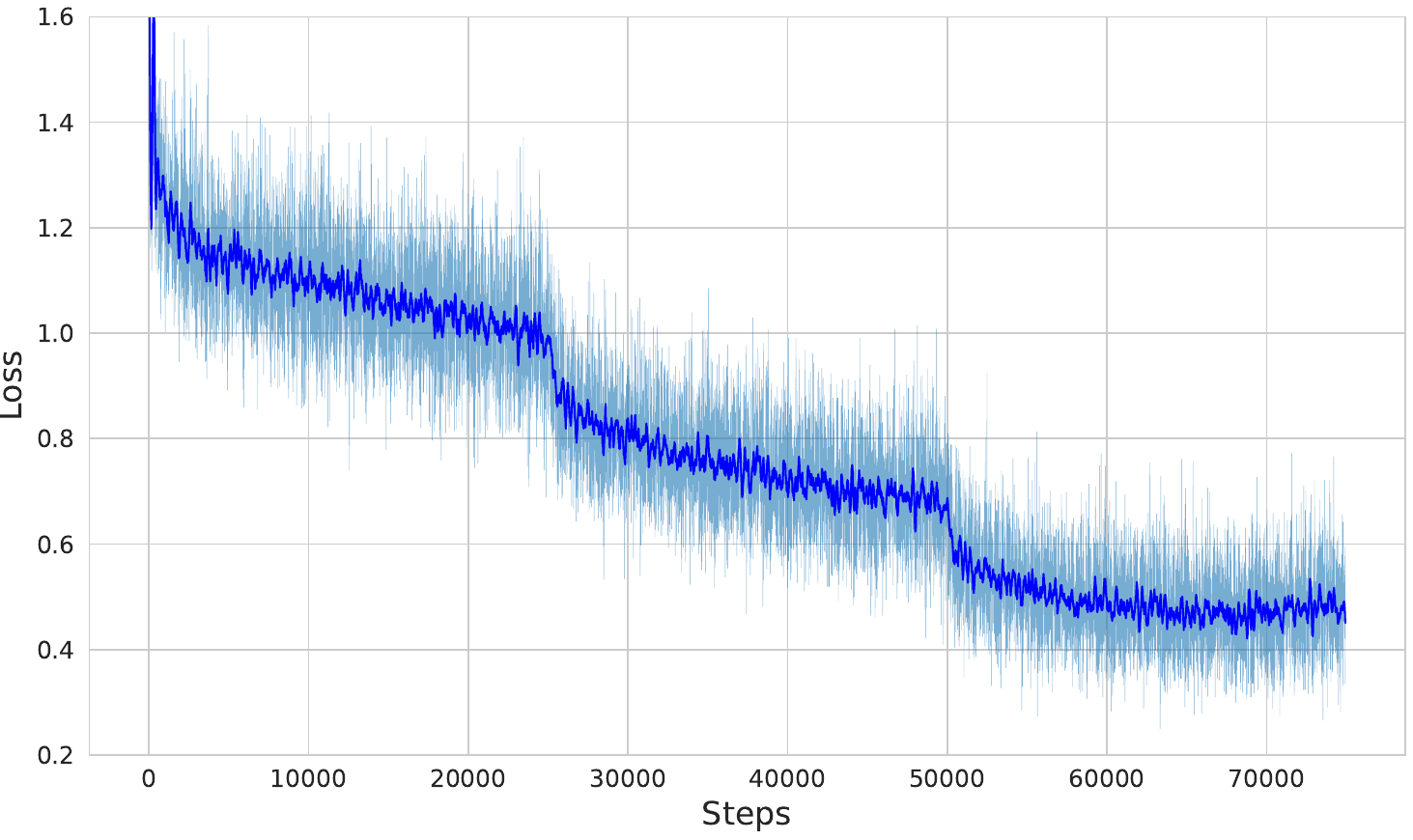}
   \vspace{-0.05in} 
  \caption{Training loss curve of BayLing-2-8B.}\label{fig:loss}
\vspace{-0.1in} 
\end{center}
\end{wrapfigure} 
Using Llama-2-7B-Chat, Llama-2-13B-Chat and Llama-3-8B-Instruct as foundational models, We fine-tune \textbf{BayLing-2-7B}, \textbf{BayLing-2-13B} and \textbf{BayLing-2-8B} respectively on the instruction dataset proposed in Section~\ref{sec:Instruction}. We fine-tune BayLing 2 on 8 NVIDIA A800 80G GPUs for 3 epochs, using a global batch size of 128, learning rate of 2e-5 and weight decay of 0.0. Note that we apply learning rate of 2e-6 for BayLing-2-8B.
We employ DeepSpeed \citep{10.1145/3394486.3406703} and Gradient Checkpointing \citep{chen2016training} techniques to optimize memory consumption. The training loss curve of BayLing-2-8B is depicted in Figure~\ref{fig:loss}.

\section{Evaluation}

In this section, we comprehensively evaluate the performance of BayLing-2-7B, BayLing-2-13B and BayLing-2-8B on multilingual tasks and general tasks respectively.

\subsection{Multilingual Capability}

BayLing's multilingual capabilities are primarily manifested in two aspects: multilingual translation and multilingual interaction. Multilingual translation aims to accomplish translation between different languages, which can be utilized to assess the language alignment within LLMs as well as the comprehension and generation capabilities across different languages. Multilingual interaction involves multitask language understanding using multiple languages, which can be employed to evaluate the multilingual knowledge and reasoning abilities of LLMs.

\subsubsection{Multilingual Translation}
We employ multilingual translation to assess the multilingual alignment within LLMs, which entails producing outputs that retain the same meaning but in different languages. We conduct evaluation on the \textbf{Flores-101} and \textbf{WMT22} benchmarks. For metrics, \texttt{BLEU} (sacrebleu) \citep{post-2018-call} and \texttt{COMET} \citep{rei-etal-2022-comet} are used to assess the quality of LLMs' translation. BLEU score measures the statistical similarity based on n-gram accuracy, COMET score measures the semantic similarity using cross-lingual pre-trained models, which is currently regarded as the most human-aligned evaluation metric for translation tasks.

\begin{table}[t]
\centering
\small
\caption{Mulitlingual translation preformance on WMT22 benchmark. X indicates other 100 languages in Flores-101, and the results are averaged over these 100 languages.}
\label{tab:flores}
\begin{tabular}{lC{0.9cm}C{0.9cm}C{0.9cm}C{0.9cm}C{0.9cm}C{0.9cm}C{0.9cm}C{0.9cm}} \toprule
\multirow{2}{*}{\textbf{Models}} & \multicolumn{2}{c}{\textbf{X$\Rightarrow$English}} & \multicolumn{2}{c}{\textbf{English$\Rightarrow$X}} & \multicolumn{2}{c}{\textbf{X$\Rightarrow$Chinese}} & \multicolumn{2}{c}{\textbf{Chinese$\Rightarrow$X}} \\ \cmidrule(lr){2-3} \cmidrule(lr){4-5}\cmidrule(lr){6-7}\cmidrule(lr){8-9}
                                 & BLEU               & $\!\!$COMET             & BLEU              & $\!\!$COMET              & BLEU              & $\!\!$COMET              & BLEU              & $\!\!$COMET              \\ \midrule
\textbf{Llama-1-7B}              & 14.07          & 60.94          & 6.93           & \textbf{49.73} & 0.93           & 40.88          & 1.85           & 44.88          \\
\textbf{BayLing-1-7B}            & \textbf{14.70} & \textbf{61.93} & \textbf{7.04}  & 49.33          & \textbf{1.58}  & \textbf{46.22} & \textbf{1.56}  & \textbf{48.78} \\\midrule
\textbf{Llama-2-7B-Chat }        & 15.39          & 63.95          & 7.45           & 50.97          & 1.75           & 47.19          & 1.57           & 45.70          \\
\textbf{BayLing-2-7B}            & \textbf{17.71} & \textbf{67.15} & \textbf{8.02}  & \textbf{52.37} & \textbf{2.70}  & \textbf{51.44} & \textbf{2.32}  & \textbf{49.37} \\\midrule
\textbf{Llama-3-8B-Instruct}     & 25.20          & 76.60          & 16.59          & 67.17          & \textbf{11.79}          & \textbf{71.91}          & 8.95           & 63.57          \\
\textbf{BayLing-2-8B}            & \textbf{26.77} & \textbf{77.03} & \textbf{17.91} & \textbf{70.88} & 11.31 & 69.43 & \textbf{10.64} & \textbf{67.86}        \\\bottomrule         
\end{tabular}
\end{table}

\begin{figure}[t]
\centering
\subfigure[Flores-101 X$\Rightarrow$English]{
\includegraphics[width=0.98\textwidth]{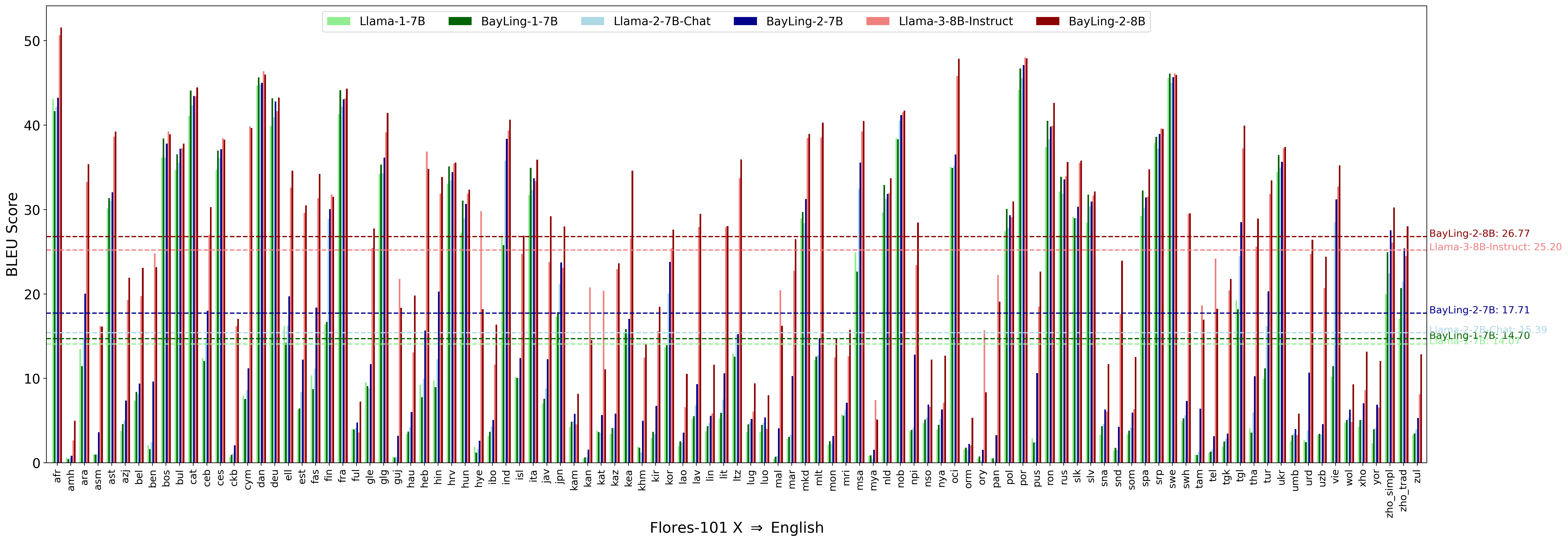}\label{fig:flores1}
}
\subfigure[Flores-101 English$\Rightarrow$X]{
\includegraphics[width=0.98\textwidth]{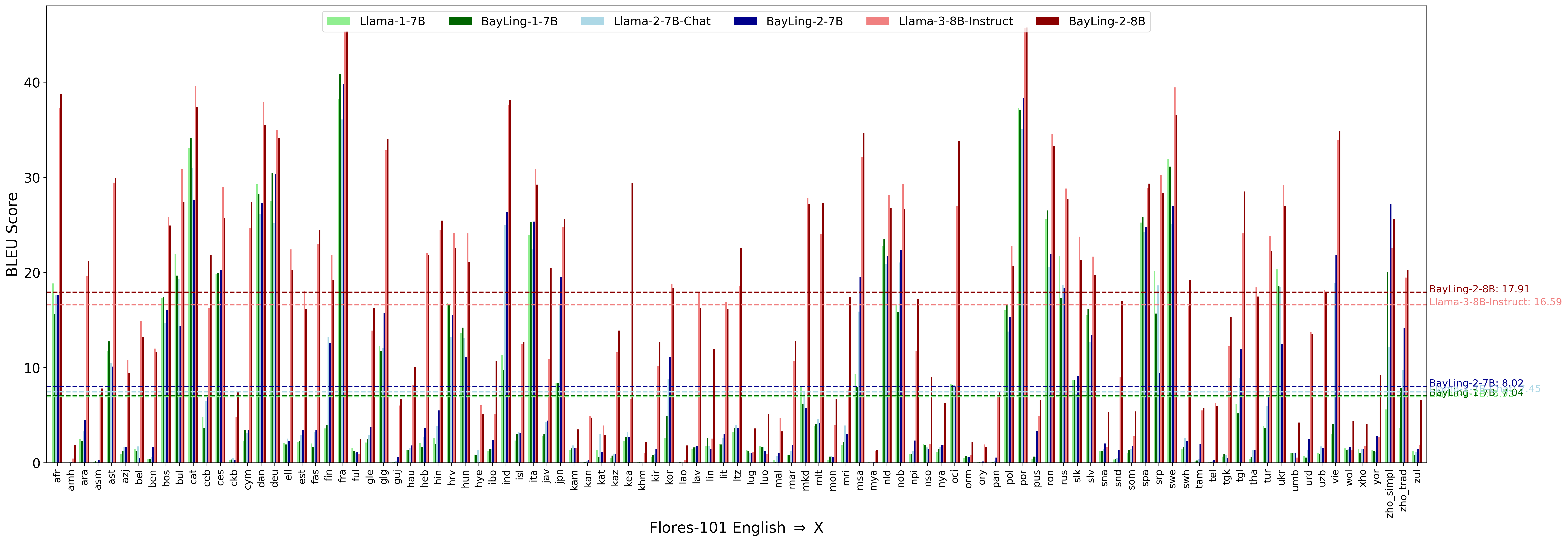}\label{fig:flores2}
}
\caption{English$\Leftrightarrow$101 languages translation performance on Flores-101 benchmark.}
\label{fig:main1}
\end{figure}

\begin{figure}[t]
\centering
\subfigure[Flores-101 X$\Rightarrow$Chinese]{
\includegraphics[width=0.98\textwidth]{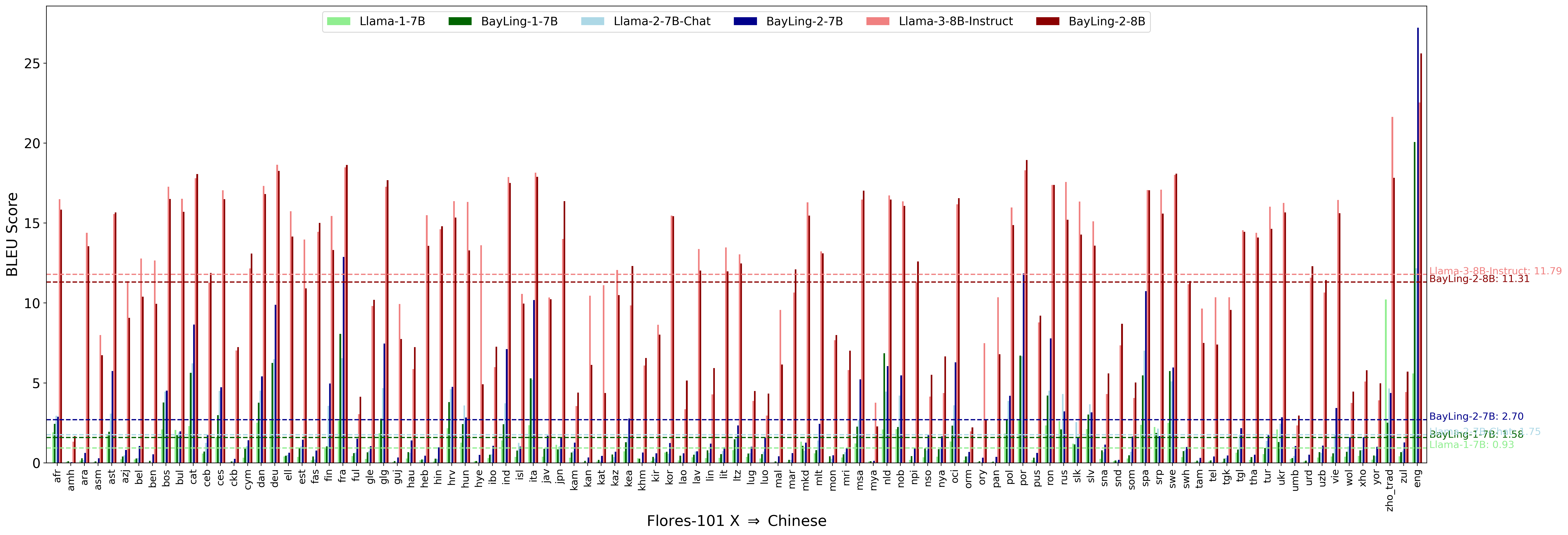}\label{fig:flores3}
}
\subfigure[Flores-101 Chinese$\Rightarrow$X]{
\includegraphics[width=0.98\textwidth]{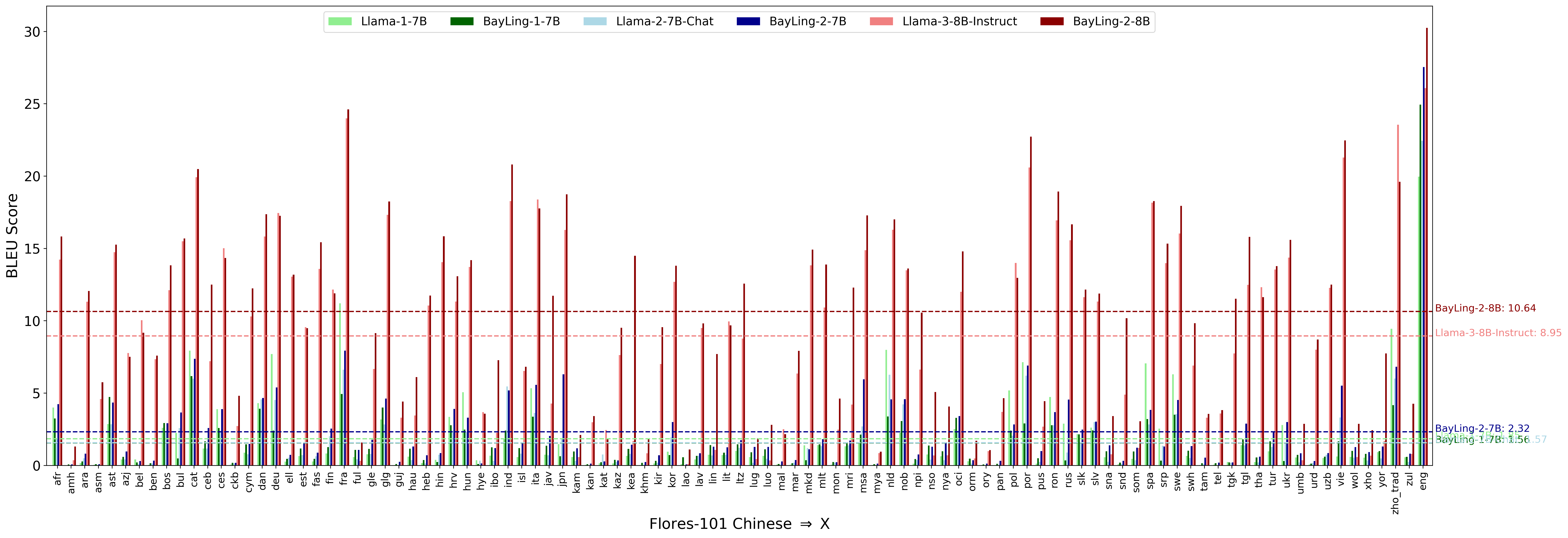}\label{fig:flores4}
}
\caption{Chinese$\Leftrightarrow$101 languages translation performance on Flores-101 benchmark.}
\label{fig:main2}
\end{figure}

\textbf{Flores-101}\quad Flores-101 benchmark encompasses 101 languages from around the world, and the sentences is sourced from various domains, including news, travel guides and books. Due to the rarity of some low-resource languages, LLMs may suffer from off-target issues. To address this, we adopt a 1-shot setting (i.e., randomly selecting an example from the dev set) to help LLMs follow the target language through in-context learning. We compare BayLing models with their corresponding foundational models, and the results are shown in Table \ref{tab:flores}. 

The results in Table \ref{tab:flores} indicate that BayLing achieves better performance in most translation directions between 100 languages and Chinese/English. Specifically, compared to the foundation LLMs Llama-1-7B and Llama-2-7B-Chat, which have relatively weak multilingual capabilities, BayLing effectively scales their language understanding and generation capabilities to over 100 languages, leading to significantly improved translation performance. Furthermore, Figures \ref{fig:flores1}, \ref{fig:flores2}, \ref{fig:flores3} and \ref{fig:flores4} illustrate the specific BLEU score improvements achieved by BayLing across 100 languages. BayLing consistently delivers the highest translation quality for most languages, particularly in translation directions to low-resource languages. This demonstrates BayLing's potential to enhance LLM in serving such low-source linguistic communities.

\begin{table}[t]
\centering\scriptsize
\caption{Mulitlingual translation preformance on WMT22 benchmark. The bold and underlined results indicate the first and second best, respectively.}
\label{tab:wmt22}
\begin{tabular}{lC{0.51cm}C{0.51cm}C{0.51cm}C{0.51cm}C{0.51cm}C{0.51cm}C{0.51cm}C{0.51cm}C{0.51cm}C{0.51cm}C{0.51cm}C{0.51cm}} \toprule
\multirow{2}{*}{\textbf{Systems}} & \multicolumn{2}{c}{\textbf{En$\Rightarrow$Zh}} & \multicolumn{2}{c}{\textbf{En$\Rightarrow$De}} & \multicolumn{2}{c}{\textbf{En$\Rightarrow$Cs}} & \multicolumn{2}{c}{\textbf{En$\Rightarrow$Ja}} & \multicolumn{2}{c}{\textbf{En$\Rightarrow$Ru}} & \multicolumn{2}{c}{\textbf{En$\Rightarrow$Uk}} \\
                                  & $\!\!\!\!$COMET            & $\!$BELU            & $\!\!\!\!$COMET            & $\!$BELU            & $\!\!\!\!$COMET            & $\!$BELU            & $\!\!\!\!$COMET            & $\!$BELU            & $\!\!\!\!$COMET            & $\!$BELU            & $\!\!\!\!$COMET            & $\!$BELU            \\ \midrule
\multicolumn{13}{c}{\textit{closed-sourced}}                                                                                                                                                                                                             \\ \midrule
\textbf{GPT-4}                             & \textbf{87.49}   & 43.98           & \textbf{87.44}   & 35.38           & 90.77            & 34.53           & 89.87            & 24.71           & 88.87            & 30.45           & 88.46            & 26.71           \\
\textbf{GPT-3.5-turbo }                    & 86.81            & 44.99           & 86.93            & 34.12           & 90.05            & 32.71           & 83.26            & 22.22           & 87.52            & 29.59           & 87.43            & 25.87           \\
\textbf{Google Translate}                  & 87.34            & \textbf{49.89}  & 87.08            & \textbf{38.27}  & \textbf{91.28}   & \textbf{48.10}  & \textbf{88.64}   & \textbf{26.50}  & \textbf{88.91}   & \textbf{35.04}  & \textbf{88.63}   & \textbf{32.05}  \\\midrule
\multicolumn{13}{c}{\textit{open-sourced}}                                                                                                                                                                                                             \\ \midrule
\textbf{Llama-2-7B-Chat}                   & 67.90            & 17.50           & 72.22            & 16.74           & 65.17            & 11.69           & 69.66            & 9.52            & 67.60            & 12.47           & 66.94            & 10.65           \\
\textbf{Llama-2-13B-Chat}                  & 75.23            & 24.31           & 77.25            & 20.35           & 75.42            & 16.18           & 78.46            & 13.56           & 77.19            & 17.11           & 75.41            & 14.75           \\
\textbf{Vicuna-7B-v1.5}                    & 81.40            & 29.54           & 75.25            & 16.65           & 71.84            & 13.63           & 74.80            & 11.28           & 77.66            & 17.95           & 74.96            & 13.26           \\
\textbf{Vicuna-13B-v1.5}                   & 84.01            & 34.69           & 81.99            & 24.22           & 77.97            & 17.47           & \textbf{85.45}            & \textbf{17.54}           & 83.31            & 21.60           & \underline{81.32}            & 17.86           \\
\textbf{Llama-3-8B-Instruct}               & 80.55            & 30.10           & 82.18            & 25.83           & \underline{83.24}            & \underline{23.41}           & 65.43            & 10.57           & 82.92            & \underline{23.53}           & 80.69            & \underline{18.88}           \\ \hline
\textbf{BayLing-1-7B}                      & 84.43            & 38.19           & 82.18            & 25.66           & 76.85            & 15.64           & 71.23            & 4.51            & 74.72            & 14.85           & 76.01            & 11.66           \\
\textbf{BayLing-1-13B}                     & 84.62            & 37.92           & 82.69            & 25.62           & 78.22            & 16.43           & 71.39            & 6.05            & 71.01            & 12.77           & 66.83            & 8.32            \\
\textbf{BayLing-2-7B}                      & \underline{85.94}            & 39.71           & 82.76            & 25.65           & 80.54            & 17.81           & \underline{84.94}   & 16.43           & 82.03            & 19.72           & 75.85            & 12.25           \\
\textbf{BayLing-2-13B}                     & \textbf{86.65}   & \textbf{42.87}  & \underline{83.79}            & \underline{26.61}           & 82.93            & 18.52           & 83.60            & 15.79           & \underline{85.23}            & 22.95           & 80.89            & 14.60           \\
\textbf{BayLing-2-8B}                      & 85.75           & \underline{41.49}           & \textbf{84.53}   & \textbf{29.59}  & \textbf{87.55}   & \textbf{24.57}  & 83.34            & \underline{16.82}  & \textbf{86.79}   & \textbf{26.41}  & \textbf{85.97}   & \textbf{21.81}  \\ \bottomrule\toprule
\multirow{2}{*}{\textbf{Systems}} & \multicolumn{2}{c}{\textbf{Zh$\Rightarrow$En}} & \multicolumn{2}{c}{\textbf{De$\Rightarrow$En}} & \multicolumn{2}{c}{\textbf{Cs$\Rightarrow$En}} & \multicolumn{2}{c}{\textbf{Ja$\Rightarrow$En}} & \multicolumn{2}{c}{\textbf{Ru$\Rightarrow$En}} & \multicolumn{2}{c}{\textbf{Uk$\Rightarrow$En}} \\ 
                                  & $\!\!\!\!$COMET            & $\!$BELU            & $\!\!\!\!$COMET            & $\!$BELU            & $\!\!\!\!$COMET            & $\!$BELU            & $\!\!\!\!$COMET            & $\!$BELU            & $\!\!\!\!$COMET            & $\!$BELU            & $\!\!\!\!$COMET            & $\!$BELU            \\ \midrule
                                  \multicolumn{13}{c}{\textit{closed-sourced}}                                                                                                                                                                                                             \\ \midrule
\textbf{GPT-4}                             & \textbf{82.79}   & 27.20           & \textbf{85.62}   & \textbf{33.87}  & 87.43            & 48.67           & 83.20            & \textbf{24.57}  & \textbf{86.18}   & \textbf{43.51}  & \textbf{85.67}   & 40.47           \\
\textbf{GPT-3.5-turbo}                     & 82.64            & 26.13           & 85.47            & 32.94           & 86.75            & 45.99           & \textbf{82.39}   & 22.14           & 85.95            & 41.79           & 85.32            & 39.00           \\
\textbf{Google Translate}                  & 80.81            & \textbf{28.63}  & 84.75            & 33.21           & \textbf{86.95}   & \textbf{49.26}  & 81.69            & 23.17           & 84.81            & 43.54           & 85.55            & \textbf{41.60}  \\\midrule
\multicolumn{13}{c}{\textit{open-sourced}}                                                                                                                                                                                                                      \\\midrule
\textbf{Llama-2-7B-Chat}                   & 75.31            & 15.42           & 80.14            & 24.53           & 78.18            & 28.91           & 74.11            & 11.38           & 80.56            & 31.23           & 79.41            & 28.68           \\
\textbf{Llama-2-13B-Chat}                  & 75.90            & 16.45           & 81.43            & 25.69           & 81.28            & 33.97           & 76.37            & 13.37           & 81.60            & 32.72           & 81.19            & 31.55           \\
\textbf{Vicuna-7B-v1.5 }                   & 75.42            & 16.80           & 79.07            & 23.57           & 76.34            & 24.46           & 72.13            & 10.89           & 78.63            & 27.95           & 78.29            & 25.74           \\
\textbf{Vicuna-13B-v1.5}                   & 78.47            & 19.41           & 83.25            & 29.19           & 81.71            & 34.51           & 75.22            & 13.66           & 82.18            & 33.74           & \underline{82.54}            & 33.03           \\
\textbf{Llama-3-8B-Instruct}               & \textbf{80.44}   & 21.57           & \textbf{83.84}   & \textbf{29.37}  & \textbf{83.47}   & \textbf{39.49}  & \underline{78.83}            & \underline{17.00}           & \textbf{84.08}   & \textbf{37.05}  & \textbf{82.75}   & \textbf{34.53}  \\\hline
\textbf{BayLing-1-7B}                      & 77.48            & 20.31           & 83.19            & 28.16           & 82.03            & 35.98           & 72.16            & 11.63           & 82.48            & \underline{34.74}           & 81.38            & 33.07           \\
\textbf{BayLing-1-13B }                    & 77.72            & 20.12           & 83.02            & 27.34           & 81.65            & 33.87           & 72.14            & 12.23           & 82.07            & 33.95           & 81.41            & 32.67           \\
\textbf{BayLing-2-7B }                     & 79.07            & 22.09           & 82.56            & 27.33           & 80.66            & 31.91           & 76.35            & 15.12           & 81.19            & 29.37           & 80.34            & 28.60           \\
\textbf{BayLing-2-13B }                    & 79.47            & \textbf{23.43}  & 83.07            & 28.81           & 81.63            & 34.08           & 76.55            & 14.95           & 81.71            & 31.75           & 80.56            & 30.09           \\
\textbf{BayLing-2-8B}                      & \underline{79.75}            & \underline{22.58}           & \underline{83.55}            & \underline{28.99}           & \underline{83.16}            & \underline{37.43}           & \textbf{79.25}   & \textbf{18.61}  & \underline{83.15}            & 34.07           & 81.89            & \underline{31.98}          \\\bottomrule
\end{tabular}
\end{table}

\textbf{WMT22}\quad WMT22 benchmark\footnote{\url{https://www.statmt.org/wmt22/translation-task.html}} encompass is used to evaluate high-resource multilingual translation performance, including translation directions of Chinese$\Leftrightarrow$English, German$\Leftrightarrow$English, Czech$\Leftrightarrow$English, Japanese$\Leftrightarrow$English, Russian$\Leftrightarrow$English, and Ukrainian$\Leftrightarrow$English. We compared BayLing with the best closed-sourced and open-sourced models, including \textbf{GPT-4}\footnote{We use GPT-4 API of version 0314} \citep{openai2023gpt4}, \textbf{GPT-3.5-turbo}\footnote{We use GPT-3.5-turbo API} \citep{chatgpt}, \textbf{Google Translate}\footnote{\url{https://translate.google.com/}}, \textbf{Llama}\citep{touvron2023llama} and \textbf{Vicuna} \citep{vicuna2023}.

The translation results on WMT22 are shown in Table \ref{tab:wmt22}, where the results illustrate the superior multilingual translation capabilities of BayLing models. Among the open-sourced models, BayLing achieves the highest overall translation performance, coming remarkably close to the performance levels of closed-sourced models like GPT-4 and GPT-3.5-turbo. This exceptional performance can be attributed to BayLing's improved language alignment, which enables it to produce more accurate and reliable translations across different languages. In particular,for the Zh$\Leftrightarrow$En translation, BayLing-2-8B achieves a COMET score of 79.75 on Zh$\Rightarrow$En and 85.75 on En$\Rightarrow$Zh, which is very close to the performance of Google Translate. 

\textbf{Improving Mulitlingual Generation Capabilities}\quad We have observed that foundational models often exhibit off-target issues when generating low-resource languages. In contrast, BayLing demonstrates significantly enhanced multilingual generation capabilities, consistently improving translation performance from English to other languages. This indicates that BayLing can activate the multilingual generation abilities of LLMs solely through cross-lingual translation data, without the need for extensive multilingual instruction data. This finding is crucial for efficiently enhancing the multilingual capabilities of LLMs, as it is nearly impossible to collect instruction data covering more than 100 languages while multilingual translation data is relatively abundant and easier to obtain. BayLing's approach of transferring generation capabilities from high-resource to low-resource languages through language alignment offers an efficient solution for enhancing the multilingual generation capabilities of LLMs.

The superior multilingual translation capabilities on Flores-101 and WMT22 underscores BayLing's potential as a leading tool in the field of multilingual translation, offering significant advancements in multilingual capabilities of LLM.

\subsubsection{Multilingual Multi-task Evaluation}\label{sec:MME}

We assessed the multilingual performance of BayLing using several benchmarks. All evaluations were conducted through the Language Model Evaluation Harness\footnote{\url{https://github.com/EleutherAI/lm-evaluation-harness}} \citep{eval-harness}, an open-source, unified framework designed to assess LLMs across a wide variety of evaluation tasks. Each result was obtained in a zero-shot setting. The models Llama-2-7B, Llama-2-7B-Chat, Llama-3-8B-Instruct, Vicuna-7B and Mistral-7B served as baselines for comparison. The multilingual benchmarks are discribed as follows.

\textbf{Belebele} \citep{bandarkar2023belebele}
Belebele is a multiple-choice machine reading comprehension benchmark, which evaluates mono- and multi-lingual models across different resource levels with rigorously checked questions. Each question has four multiple-choice answers and is linked to a short passage from the FLORES-200 dataset.

\textbf{Multilingual HellaSwag} \citep{dac2023okapi}
Multilingual HellaSwag is a multilingual adaptation of HellaSwag, a benchmark dataset designed to assess commonsense inference. Despite its questions being straightforward for humans, state-of-the-art models struggle with it, highlighting the challenges in AI comprehension.

\textbf{XNLI} \citep{conneau2018xnli}
XNLI is an evaluation dataset created by extending the MultiNLI corpus to multiple languages, including low-resource ones like Swahili and Urdu. It serves as a standardized benchmark for assessing cross-lingual sentence understanding, aiming to foster research in this area.

\textbf{Multilingual ARC} \citep{dac2023okapi}
The Multilingual ARC, a multilingual extension of ARC \citep{Clark2018ThinkYH}, encompasses science examination queries, stratified into a Challenge Set comprising intricate questions and an Easy Set. All queries adhere to a multiple-choice structure.

\begin{figure}[t]
\centering
\subfigure[Belebele]{
\includegraphics[width=0.48\textwidth]{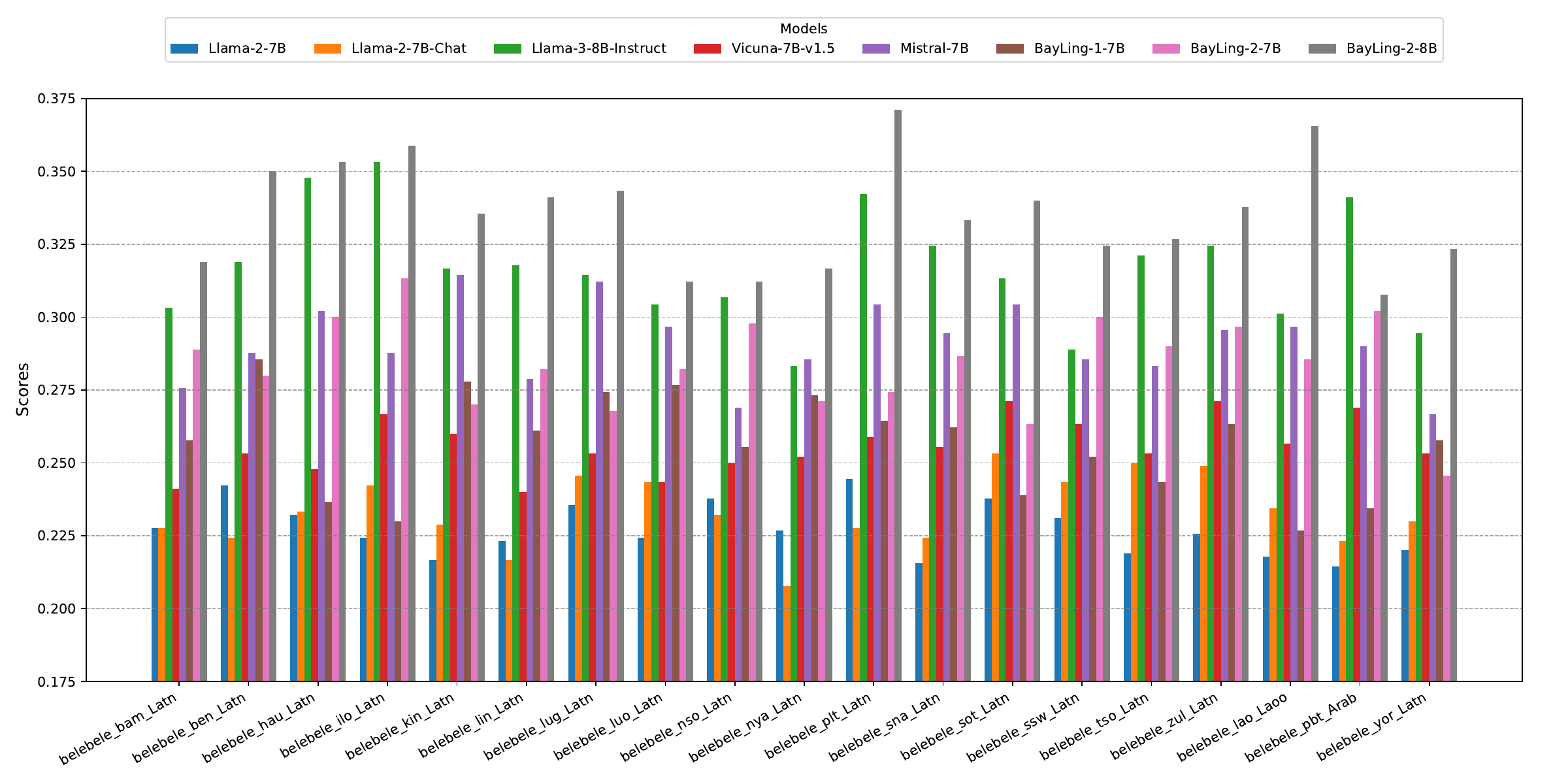}\label{fig:bel}
}
\subfigure[Multilingual HellaSwag]{
\includegraphics[width=0.48\textwidth]{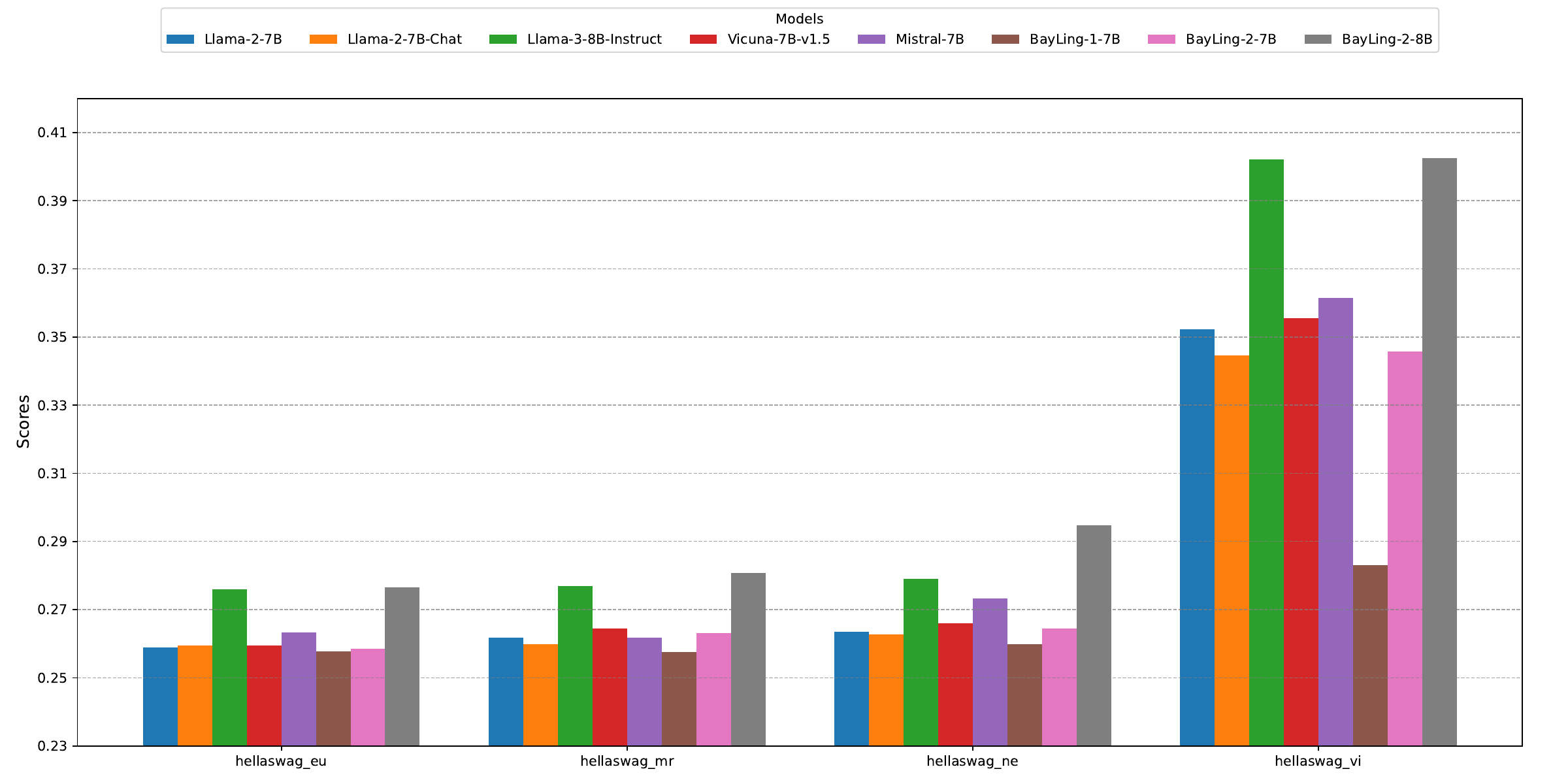}\label{fig:hel}
}
\subfigure[XNLI]{
\includegraphics[width=0.48\textwidth]{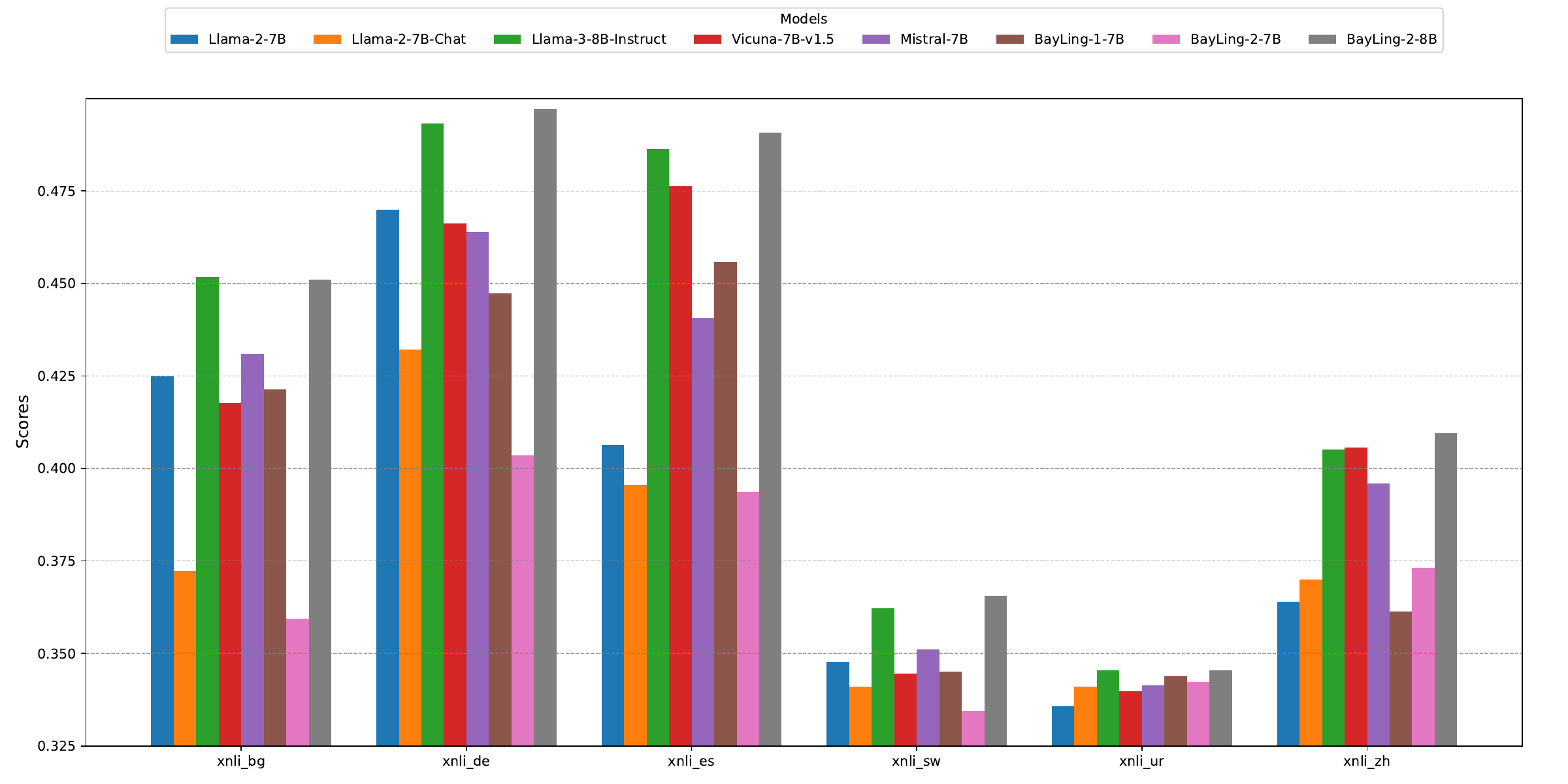}\label{fig:xnl}
}
\subfigure[Multilingual ARC]{
\includegraphics[width=0.48\textwidth]{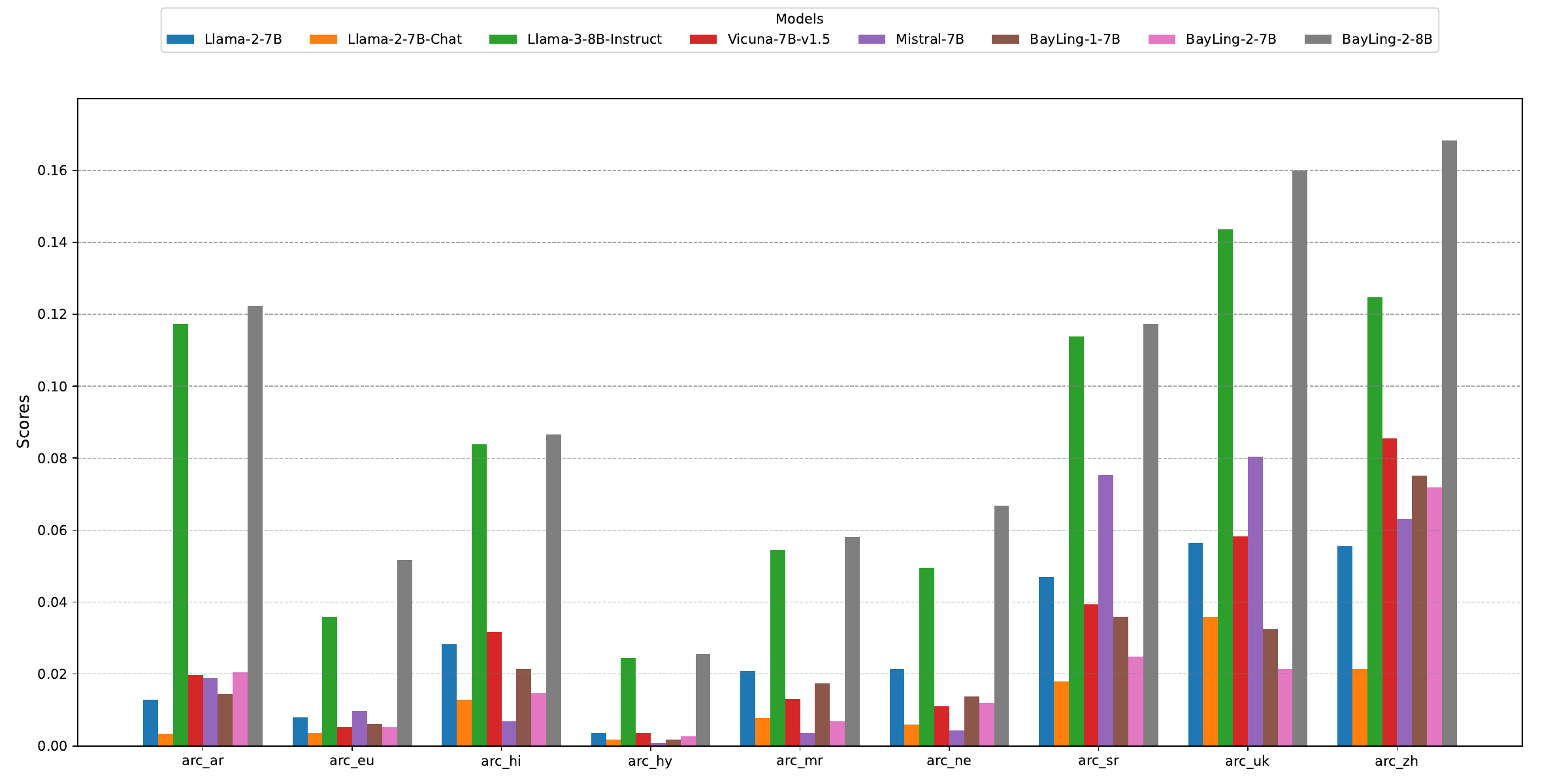}\label{fig:arc}
}

\caption{Multilingual multi-task performance of BayLing on low-resource languages.}
\label{fig:multi}
\end{figure}

Figure \ref{fig:bel}, \ref{fig:hel}, \ref{fig:xnl}, \ref{fig:arc} provide detailed illustrations of the experimental outcomes on the Belebele, Multilingual HellaSwag, XNLI, Multilingual ARC benchmarks across several low-resource languages. The BayLing-2-7B and BayLing-2-8B models demonstrate notable performance benefits. Among these, BayLing-2-8B consistently delivers the best results across most of the low-resource languages evaluated. Meanwhile, BayLing-2-7B outperforms other 7B models in most of these languages. Remarkably, BayLing-2-7B even surpasses the Llama-3-8B-Instruct model in the Swati language subset on Belebele benchmark (belebele\_ssw\_Latn). 

Note that BayLing's training data does not include instruction data for these low-resource languages but only cross-lingual instructions between these low-resource languages and Chinese/English. Therefore, the performance improvements observed in these low-resource languages demonstrate that BayLing effectively transfers knowledge and understanding capability from high-resource languages to low-resource ones through language alignment. Overall, BayLing's approach of leveraging language alignment for capability transfer offers an efficient solution to enhance LLM's performance in low-resource languages.

\subsection{General Capability}

Furthermore, we evaluated the general capability of BayLing on the following benchmarks employing the same settings as described in section \ref{sec:MME}.

\textbf{CMMLU} \citep{li2023cmmlu}
CMMLU serves as a specialized evaluation benchmark tailored to assess the knowledge and reasoning capacities of LLMs in within the context of Chinese language and culture. Encompassing a wide range of subjects, CMMLU includes 67 topics, which vary from basic to advanced professional levels.

\textbf{C-Eval} \citep{huang2023ceval}
C-Eval is an exhaustive Chinese evaluation suite designed for foundation models. It features a total of 13,948 multiple-choice questions, covering 52 distinct disciplines across four levels of difficulty.

\textbf{Arabic EXAMS} \citep{hardalov2020exams}
The Arabic EXAMS comprise the Arabic segment of EXAMS, a resource dedicated to multilingual high school examination questions. This section includes five subjects: Islamic Studies, Biology, Physics, General Science, and Social Studies.

\textbf{ANLI} \citep{nie-etal-2020-adversarial}
Adversarial NLI (ANLI) is a dataset assembled through an iterative adversarial procedure involving both human and model participation. It is structured into three rounds, each escalating in difficulty and complexity. Additionally, each question-answer pair in the dataset is supplemented with explanations provided by the annotators.

\textbf{CB} \citep{de2019commitmentbank, NEURIPS2019_4496bf24}
CB (CommitmentBank) is a corpus featuring texts with embedded clauses evaluated for the author's commitment to their truth. This corpus is used in a three-class textual entailment task. Examples are organized with a premise and a corresponding hypothesis extracted from the embedded clause.

\textbf{GLUE} \citep{wang-etal-2018-glue}
The GLUE benchmark is a benchmark for evaluating natural language understanding systems. It consists of nine language understanding tasks and a diagnostic dataset for assessing model performance across linguistic phenomena.

\textbf{ACLUE} \citep{zhang-li-2023-large}
ACLUE is a benchmark designed to evaluate large language models' comprehension of ancient Chinese, featuring 15 tasks across multiple domains. The questions, covering historical periods from the Xia to the Ming dynasty, are presented in a multiple-choice format.

\textbf{GSM8K} \citep{cobbe2021training}
GSM8K is a benchmark used to evaluate the math capability of LLMs as described in section \ref{sec:MME}.

\begin{figure}[t]
    \centering
    \includegraphics[width=0.99\textwidth]{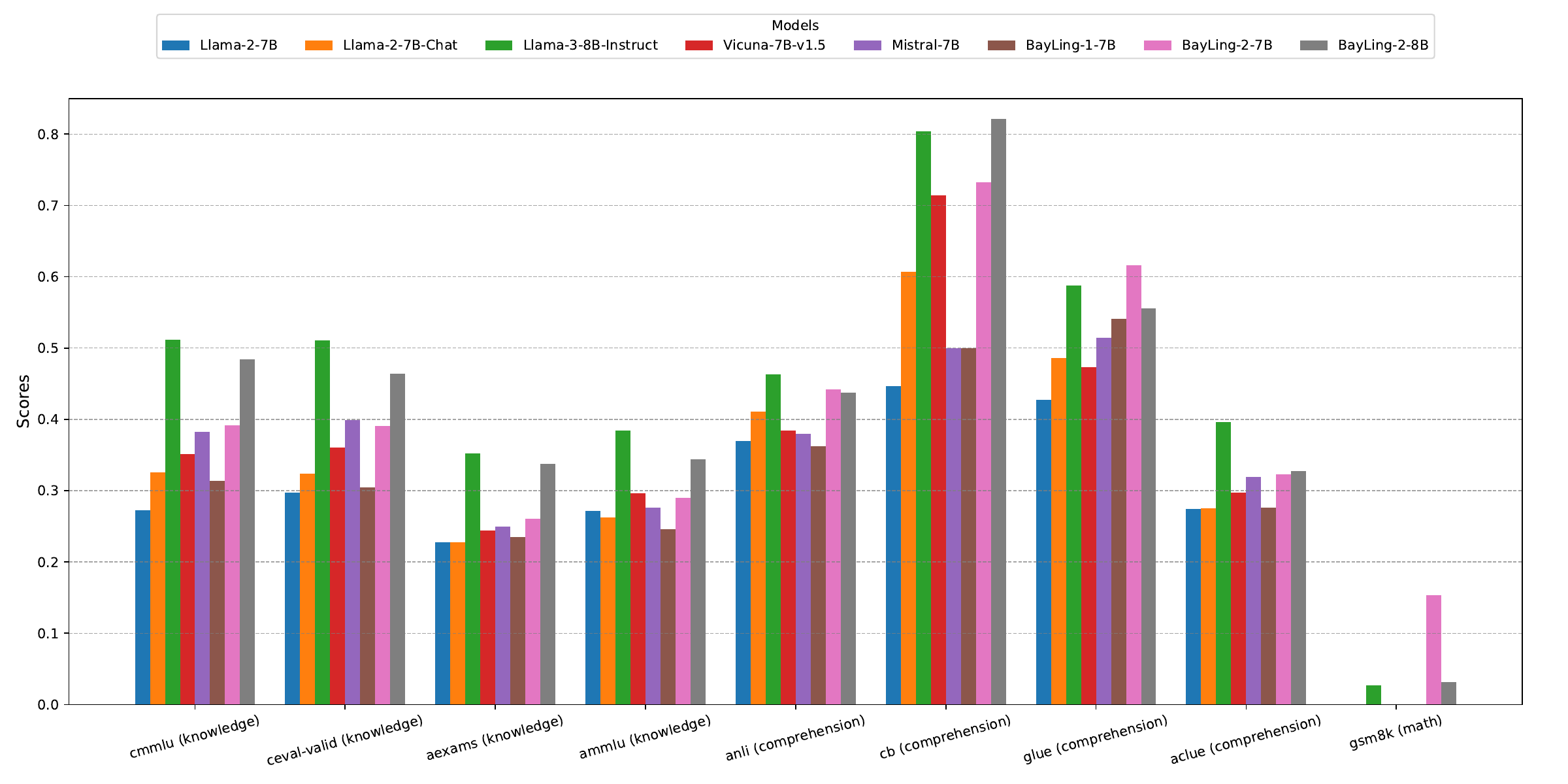}
    \caption{Scores on general benchmarks.}
    \label{fig:gen}
\end{figure}

Figure \ref{fig:gen} illustrates the performance of BayLing on the specified benchmarks. BayLing-2-7B and BayLing-2-8B demonstrate exceptional performance across several benchmarks. Notably, BayLing-2-8B outperforms all other models, achieving a score of 0.8214 on the CommitmentBank Benchmark. BayLing-2-7B attains the highest performance on GLUE and GSM8K benchmarks. Despite not being specifically trained for it, BayLing-2-7B and BayLing-2-8B still deliver comparable performances on other benchmarks when compared to other models. Overall, BayLing enhances the multilingual capabilities of LLMs, especially in low-resource languages, without significantly impacting the performance in high-resource languages. This indicates that BayLing effectively mitigates multilingual conflicts within LLM through language alignment.

\begin{figure}[t]
    \centering
    \includegraphics[width=0.99\textwidth]{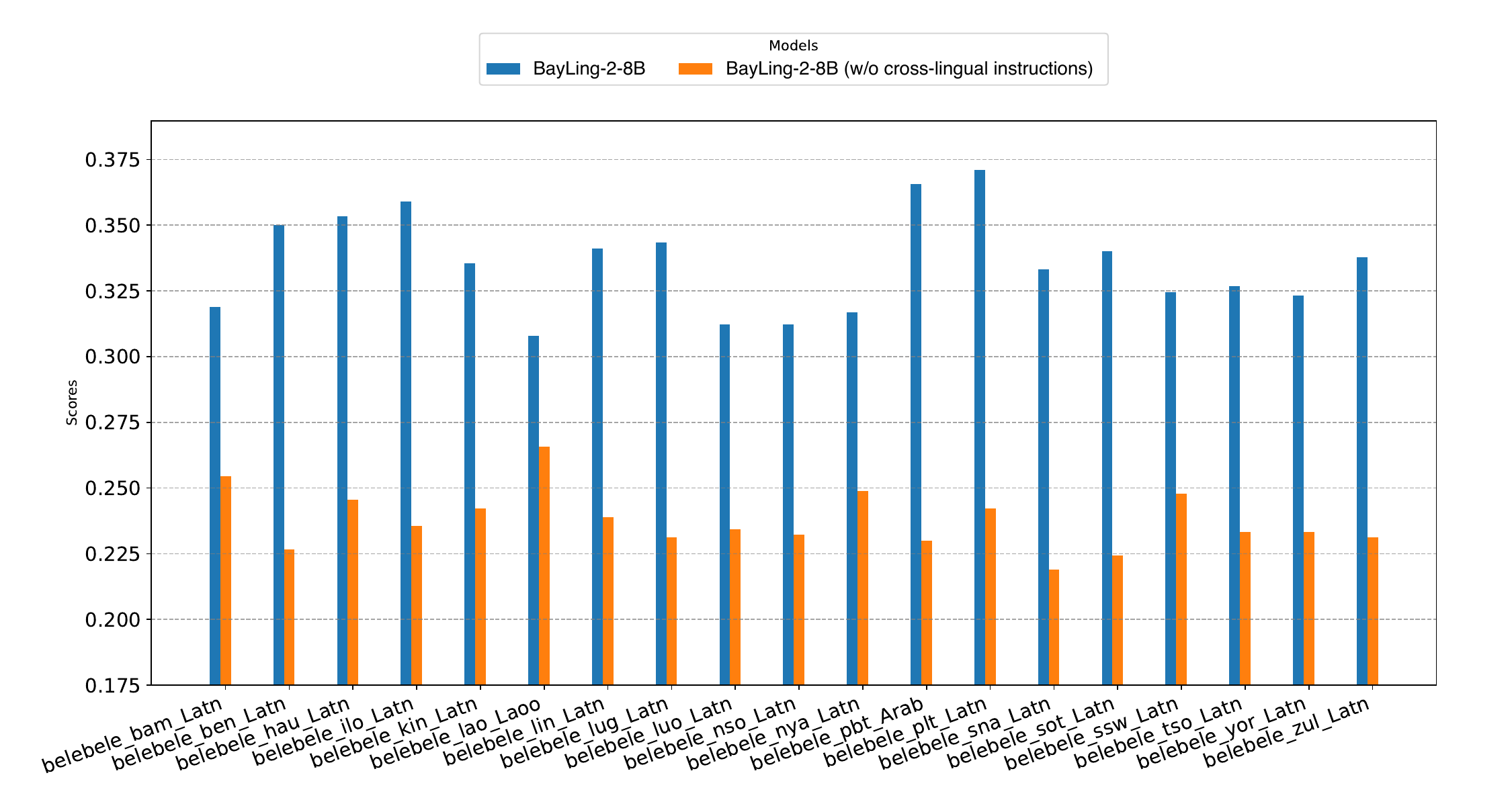}
    \caption{Effect of language alignment on multilingual benchmark Belebele.}
    \label{fig:abl_bel}
\end{figure}

\begin{figure}[t]
    \centering
    \includegraphics[width=0.99\textwidth]{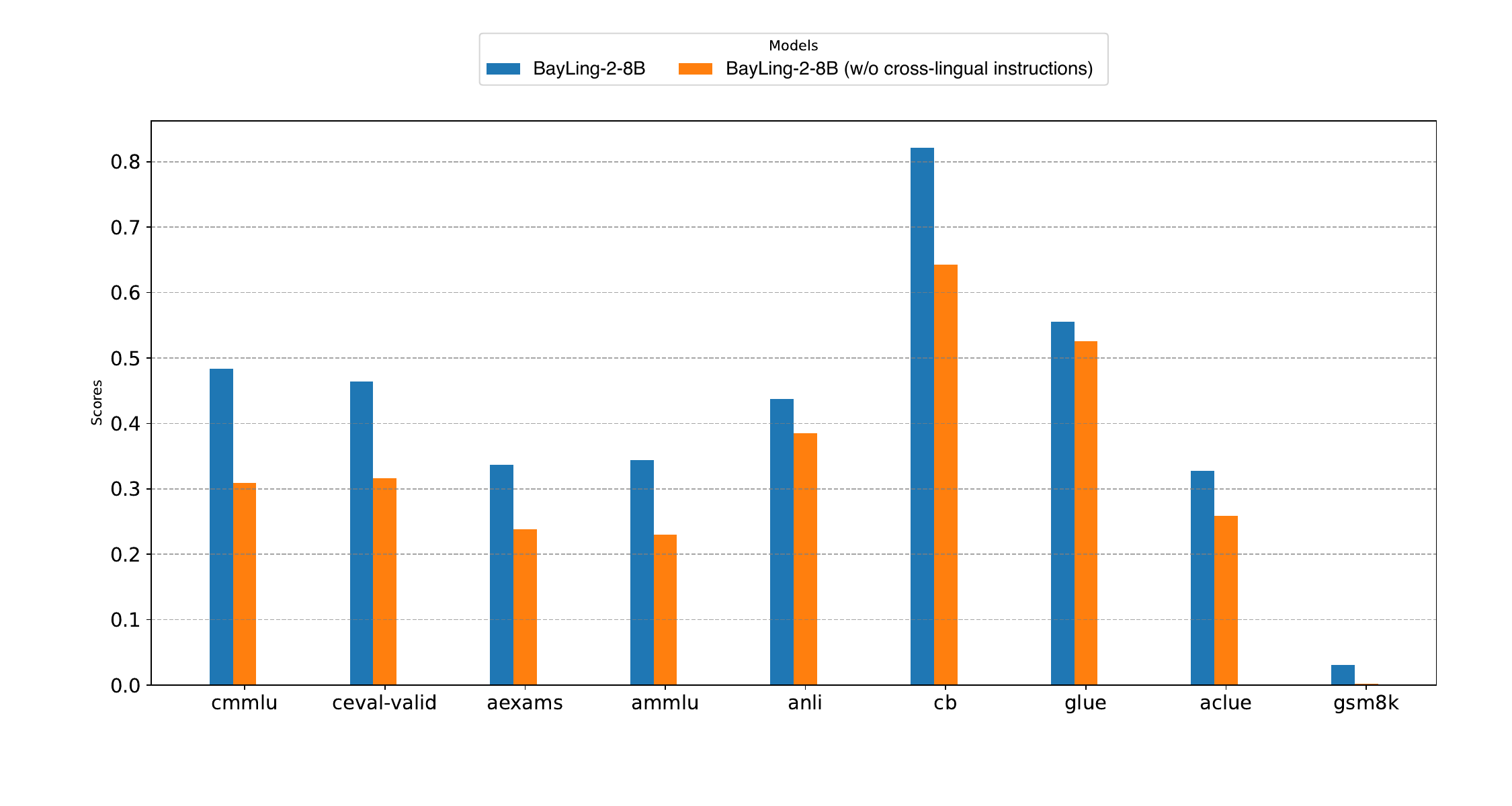}
    \caption{Effect of language alignment on Chinese and English general tasks.}
    \label{fig:abl_gen}
\end{figure}

\subsection{Effect of Language Alignment}

To validate the effect of language alignment brought by cross-lingual instructions, we conducted an ablation study on cross-lingual instructions. Specifically, we removed all cross-lingual instructions from the training data, denoting this variant as BayLing-2-8B (w/o cross-lingual instructions).

\textbf{Improving Performance of Low-resource Languages}\quad Figure \ref{fig:abl_bel} compares the performance of BayLing-2-8B and BayLing-2-8B (w/o cross-lingual instructions) on the multilingual benchmark Belebele. The results show that cross-lingual instructions significantly enhance LLM performance in low-resource languages. This indicates that cross-lingual instructions successfully help LLM achieve language alignment, thereby transferring knowledge and comprehension from high-resource languages to low-resource ones. When removing all cross-lingual instructions, the performance of LLMs in low-resource languages is adversely affected due to catastrophic forgetting. Therefore, involving cross-lingual instructions in supervised fine-tuning is both efficient and crucial for improving the multilingual capabilities of LLMs.

\textbf{Avoiding Inter-language Conflicts}\quad Figure \ref{fig:abl_gen} compares BayLing-2-8B and BayLing-2-8B (w/o cross-lingual instructions) on the Chinese/English benchmark. When removing cross-lingual instructions, we observed a significant performance decline in Chinese benchmark, indicating LLM will suffer from conflicts between Chinese and English instructions. The presence of numerous cross-lingual instructions between Chinese and English largely prevents these conflicts. Therefore, to simultaneously enhance LLM performance across multiple languages, introducing cross-lingual instructions is an effective way to avoid inter-language conflicts.

\section{Conclusion}

In this study, we develop BayLing 2, which enhances LLM's multilingual capabilities through language alignment. Adhering to an efficiency-focused approach, BayLing 2 transfers knowledge and generative abilities from high-resource languages to low-resource languages within LLM via language alignment. Comprehensive evaluation results demonstrate that BayLing 2 achieves outstanding translation performance across over 100 languages, possesses superior multilingual knowledge and understanding capability, and maintains robust proficiency in high-resource languages of Chinese and English.

\section*{Acknowledgements}

We extend our heartfelt gratitude to everyone who contributed to the development of BayLing 2. In particular, we would like to thank Ms. Xiaohong Wang for her insightful feedback and valuable suggestions regarding the use of OneAiNexus, as well as for her exceptional support in organizing resources, providing computational infrastructure, and facilitating the presentation of BayLing 2. Besides, we would like to express thanks to the Sothis.AI for the support in training of BayLing 2. Special thanks are due to the team of Nanjing Institute of InforSuperBahn - Intelligent Computility Platform Research Center, who played an indispensable role in maintaining the computational resources, designing the BayLing 2's webpage and demonstrating the system interface.

\bibliographystyle{unsrtnat}
\bibliography{customs}

\newpage
\appendix

\section{Flores-101 Benchmark}
Table \ref{tab:num_flores1}, \ref{tab:num_flores2}, \ref{tab:num_flores3}, and \ref{tab:num_flores4} report the numerical results of Llama-3-8B-Instruct and BayLing-2-8B on the Flores-101 benchmark.

\begin{table}[h]
\centering\tiny
\caption{BLEU scores of Llama-3-8B-Instruct on Flores-101 benchmark.}
\label{tab:num_flores1}
\begin{tabular} 
{llC{0.5cm}C{0.5cm}C{0.5cm}C{0.5cm}|llC{0.5cm}C{0.5cm}C{0.5cm}C{0.5cm}}\toprule
\multicolumn{12}{c}{\textbf{Llama-3-8B-Instruct}}   \\\toprule
\textbf{X}   & \textbf{Language} & \textbf{$\!\!$X$\Rightarrow$En} & \textbf{$\!\!$En$\Rightarrow$X} & \textbf{$\!\!$X$\Rightarrow$Zh} & \textbf{$\!\!$Zh$\Rightarrow$X} & \textbf{X}   & \textbf{Language} & \textbf{$\!\!$X$\Rightarrow$En} & \textbf{$\!\!$En$\Rightarrow$X} & \textbf{$\!\!$X$\Rightarrow$Zh} & \textbf{$\!\!$Zh$\Rightarrow$X}  \\ \midrule
\textbf{afr} & Afrikaans         & 50.65  & 37.33    & 16.48  & 14.22    & \textbf{lug}        & Ganda               & 6.08   & 1.11     & 3.87   & 0.44     \\
\textbf{amh} & Amharic           & 2.63   & 0.44     & 1.33   & 0.35     & \textbf{luo}        & Luo                 & 4.01   & 0.89     & 2.95   & 0.35     \\
\textbf{ara} & Arabic            & 33.28  & 19.63    & 14.38  & 11.31    & \textbf{mal}        & Malayalam           & 20.44  & 4.72     & 9.56   & 2.51     \\
\textbf{asm} & Assamese          & 16.17  & 7.20     & 7.99   & 4.60     & \textbf{mar}        & Marathi             & 22.72  & 10.64    & 10.65  & 6.35     \\
\textbf{ast} & Asturian          & 38.63  & 29.44    & 15.56  & 14.74    & \textbf{mkd}        & Macedonian          & 38.47  & 27.85    & 16.29  & 13.82    \\
\textbf{azj} & North Azerbaijani & 19.25  & 10.84    & 11.36  & 7.77     & \textbf{mlt}        & Maltese             & 38.54  & 24.08    & 13.22  & 10.92    \\
\bottomrule                          
\end{tabular}
\end{table}

\begin{table}[h]
\centering\tiny
\caption{BLEU scores of BayLing-2-8B on Flores-101 benchmark.}
\label{tab:num_flores2}
\begin{tabular} 
{llC{0.5cm}C{0.5cm}C{0.5cm}C{0.5cm}|llC{0.5cm}C{0.5cm}C{0.5cm}C{0.5cm}}\toprule
\multicolumn{12}{c}{\textbf{BayLing-2-8B}}   \\\toprule
\textbf{X}   & \textbf{Language} & \textbf{$\!\!$X$\Rightarrow$En} & \textbf{$\!\!$En$\Rightarrow$X} & \textbf{$\!\!$X$\Rightarrow$Zh} & \textbf{$\!\!$Zh$\Rightarrow$X} & \textbf{X}   & \textbf{Language} & \textbf{$\!\!$X$\Rightarrow$En} & \textbf{$\!\!$En$\Rightarrow$X} & \textbf{$\!\!$X$\Rightarrow$Zh} & \textbf{$\!\!$Zh$\Rightarrow$X}  \\ \midrule
\textbf{afr} & Afrikaans         & 51.55           & 38.76             & 15.83           & 15.83             & \textbf{lug}        & Ganda               & 9.41                                & 3.59                                  & 4.49                                & 1.88                                  \\
\textbf{amh} & Amharic           & 4.97            & 1.87              & 1.67            & 1.31              & \textbf{luo}        & Luo                 & 8.00                                & 5.15                                  & 4.33                                & 2.80                                  \\
\textbf{ara} & Arabic            & 35.36           & 21.19             & 13.54           & 12.06             & \textbf{mal}        & Malayalam           & 16.22                               & 3.29                                  & 6.15                                & 2.15                                  \\
\textbf{asm} & Assamese          & 16.12           & 7.79              & 6.72            & 5.75              & \textbf{mar}        & Marathi             & 26.48                               & 12.81                                 & 12.10                               & 7.92                                  \\
\bottomrule                          
\end{tabular}
\end{table}

\begin{table}[h]
\centering\tiny
\caption{COMET scores of Llama-3-8B-Instruct on Flores-101 benchmark.}
\label{tab:num_flores3}
\begin{tabular} 
{llC{0.5cm}C{0.5cm}C{0.5cm}C{0.5cm}|llC{0.5cm}C{0.5cm}C{0.5cm}C{0.5cm}}\toprule
\multicolumn{12}{c}{\textbf{Llama-3-8B-Instruct}}   \\\toprule
\textbf{X}   & \textbf{Language} & \textbf{$\!\!$X$\Rightarrow$En} & \textbf{$\!\!$En$\Rightarrow$X} & \textbf{$\!\!$X$\Rightarrow$Zh} & \textbf{$\!\!$Zh$\Rightarrow$X} & \textbf{X}   & \textbf{Language} & \textbf{$\!\!$X$\Rightarrow$En} & \textbf{$\!\!$En$\Rightarrow$X} & \textbf{$\!\!$X$\Rightarrow$Zh} & \textbf{$\!\!$Zh$\Rightarrow$X}  \\ \midrule
\textbf{afr} & Afrikaans         & 85.79  & 82.75    & 79.07  & 76.71    & \textbf{lug}        & Ganda               & 49.86  & 39.62    & 48.30  & 40.12    \\
\textbf{amh} & Amharic           & 57.47  & 38.74    & 50.99  & 35.71    & \textbf{luo}        & Luo                 & 46.09  & 34.60    & 45.78  & 36.63    \\
\textbf{ara} & Arabic            & 84.85  & 77.90    & 79.64  & 75.42    & \textbf{mal}        & Malayalam           & 82.12  & 54.48    & 74.40  & 49.07    \\
\textbf{asm} & Assamese          & 78.26  & 62.26    & 72.23  & 56.98    & \textbf{mar}        & Marathi             & 82.20  & 56.38    & 75.18  & 49.97    \\
\textbf{ast} & Asturian          & 80.99  & 70.31    & 78.66  & 68.53    & \textbf{mkd}        & Macedonian          & 85.40  & 82.46    & 79.96  & 77.19    \\
\textbf{azj} & North Azerbaijani & 82.72  & 74.79    & 76.83  & 68.66    & \textbf{mlt}        & Maltese             & 75.13  & 62.43    & 68.83  & 59.95    \\
\bottomrule                          
\end{tabular}
\end{table}

\begin{table}[h]
\centering\tiny
\caption{COMET scores of BayLing-2-8B on Flores-101 benchmark.}
\label{tab:num_flores4}
\begin{tabular} 
{llC{0.5cm}C{0.5cm}C{0.5cm}C{0.5cm}|llC{0.5cm}C{0.5cm}C{0.5cm}C{0.5cm}}\toprule
\multicolumn{12}{c}{\textbf{BayLing-2-8B}}   \\\toprule
\textbf{X}   & \textbf{Language} & \textbf{$\!\!$X$\Rightarrow$En} & \textbf{$\!\!$En$\Rightarrow$X} & \textbf{$\!\!$X$\Rightarrow$Zh} & \textbf{$\!\!$Zh$\Rightarrow$X} & \textbf{X}   & \textbf{Language} & \textbf{$\!\!$X$\Rightarrow$En} & \textbf{$\!\!$En$\Rightarrow$X} & \textbf{$\!\!$X$\Rightarrow$Zh} & \textbf{$\!\!$Zh$\Rightarrow$X}  \\ \midrule
\textbf{afr} & Afrikaans         & 87.38  & 82.77    & 76.93  & 76.33    & \textbf{lug}        & Ganda               & 52.34  & 54.41    & 47.80  & 55.35    \\
\textbf{amh} & Amharic           & 58.46  & 49.90    & 49.78  & 46.53    & \textbf{luo}        & Luo                 & 49.37  & 53.25    & 47.44  & 53.08    \\
\textbf{ara} & Arabic            & 85.49  & 81.47    & 76.61  & 77.17    & \textbf{mal}        & Malayalam           & 76.81  & 53.88    & 64.05  & 49.89    \\
\textbf{asm} & Assamese          & 75.40  & 61.88    & 66.47  & 57.93    & \textbf{mar}        & Marathi             & 84.00  & 61.50    & 74.95  & 55.19    \\
\textbf{ast} & Asturian          & 82.33  & 68.74    & 76.10  & 65.80    & \textbf{mkd}        & Macedonian          & 85.77  & 83.44    & 77.39  & 77.94    \\
\textbf{azj} & North Azerbaijani & 83.50  & 71.42    & 71.40  & 68.93    & \textbf{mlt}        & Maltese             & 76.80  & 63.82    & 68.15  & 60.42    \\
\bottomrule                          
\end{tabular}
\end{table}

\section{Language Code of Multilingual Benchmarks}
\label{app:lc}

Table \ref{tab:lan_code} reports the language codes for the low-resource languages in Figure \ref{fig:multi}.

\begin{table}[h]
\centering\small
\caption{Language code of Figure \ref{fig:multi}.}
\label{tab:lan_code}
\begin{tabular}{cc}\toprule
\textbf{Language Code} & \textbf{Language} \\
\midrule
belebele\_bam\_Latn & Bambara \\
belebele\_ben\_Latn & Bengali \\
belebele\_hau\_Latn & Hausa \\
belebele\_ilo\_Latn & Ilocano \\
belebele\_kin\_Latn & Kinyarwanda \\
belebele\_lao\_Laoo & Lao \\
belebele\_lin\_Latn & Lingala \\
belebele\_lug\_Latn & Luganda \\
belebele\_luo\_Latn & Luo \\
belebele\_nso\_Latn & Northern Sotho \\
belebele\_nya\_Latn & Chichewa (Nyanja) \\
belebele\_pbt\_Arab & Pashto \\
belebele\_plt\_Latn & Plateau Malagasy \\
belebele\_sna\_Latn & Shona \\
belebele\_sot\_Latn & Southern Sotho \\
belebele\_ssw\_Latn & Swazi \\
belebele\_tso\_Latn & Tsonga \\
belebele\_yor\_Latn & Yoruba \\
belebele\_zul\_Latn & Zulu \\
hellaswag\_eu & Basque (Euskara) \\
hellaswag\_mr & Marathi \\
hellaswag\_ne & Nepali \\
hellaswag\_vi & Vietnamese \\
xnli\_bg & Bulgarian \\
xnli\_de & German \\
xnli\_es & Spanish \\
xnli\_sw & Swahili \\
xnli\_ur & Urdu \\
xnli\_zh & Chinese \\
arc\_ar & Arabic \\
arc\_eu & Basque (Euskara) \\
arc\_hi & Hindi \\
arc\_hy & Armenian \\
arc\_mr & Marathi \\
arc\_ne & Nepali \\
arc\_sr & Serbian \\
arc\_uk & Ukrainian \\
arc\_zh & Chinese \\
\bottomrule
\end{tabular}
\end{table}

\section{Numerical Results of General Capability}
\label{app:num}

Table \ref{tab:model_perf} reports the numerical results of general capability in Figure \ref{fig:gen}. Table \ref{tab:model_bel}, \ref{tab:model_hel}, \ref{tab:model_xnl}, and \ref{tab:model_arc} report the numerical results of multilingual benchmarks in Figure \ref{fig:multi}. 

\begin{table}[h]
\centering\tiny
\caption{Numerical results of general capability benchmarks.}
\label{tab:model_perf}
\begin{tabular}{lccccc}\toprule
\multirow{2}{*}{\textbf{Model}} & \textbf{cmmlu} & \textbf{ceval-valid} & \textbf{aexams} & \textbf{ammlu} & \textbf{anli} \\ 
 & \textbf{(knowledge)} & \textbf{(knowledge)} & \textbf{(knowledge)} & \textbf{(knowledge)} & \textbf{(comprehension)} \\
\midrule
\textbf{Llama-2-7B}          & 0.2726           & 0.2972                 & 0.2272            & 0.2717           & 0.3697              \\
\textbf{Llama-2-7B-Chat}     & 0.3257           & 0.3239                 & 0.2272            & 0.2624           & 0.4106              \\
\textbf{Llama-3-8B-Instruct} & 0.5120           & 0.5111                 & 0.3520            & 0.3838           & 0.4634              \\
\textbf{Vicuna-7B-v1.5}      & 0.3514           & 0.3603                 & 0.2439            & 0.2962           & 0.3847              \\
\textbf{Mistral-7B}          & 0.3825           & 0.3990                 & 0.2495            & 0.2763           & 0.3800              \\
\textbf{BayLing-1-7B}        & 0.3136           & 0.3046                 & 0.2346            & 0.2458           & 0.3622              \\
\bottomrule
\end{tabular}
\end{table}

\begin{table}[h]
\centering\tiny
\caption{Numerical results of Belebele benchmark.}
\label{tab:model_bel}
\begin{tabular}{lccccc}\toprule
\textbf{Model}               & belebele\_bam\_Latn & belebele\_ben\_Latn & belebele\_hau\_Latn & belebele\_ilo\_Latn & belebele\_kin\_Latn \\
\midrule
\textbf{Llama-2-7B}          & 0.2278              & 0.2422              & 0.2322              & 0.2244              & 0.2167              \\
\textbf{Llama-2-7B-Chat}     & 0.2278              & 0.2244              & 0.2333              & 0.2422              & 0.2289              \\
\textbf{Llama-3-8B-Instruct} & 0.3033              & 0.3189              & 0.3478              & 0.3533              & 0.3167              \\
\textbf{Vicuna-7B-v1.5}      & 0.2411              & 0.2533              & 0.2478              & 0.2667              & 0.2600              \\
\textbf{Mistral-7B}          & 0.2756              & 0.2878              & 0.3022              & 0.2878              & 0.3144              \\
\textbf{BayLing-1-7B}        & 0.2578              & 0.2856              & 0.2367              & 0.2300              & 0.2778              \\
\textbf{BayLing-2-7B}        & 0.2889              & 0.2800              & 0.3000              & 0.3133              & 0.2700              \\
\bottomrule
\end{tabular}%
\end{table}
\begin{table}[h]
\centering\tiny
\caption{Numerical results of multilingual HellaSwag benchmark.}
\label{tab:model_hel}
\begin{tabular}{lcccc}\toprule
\textbf{Model}               & hellaswag\_eu & hellaswag\_mr & hellaswag\_ne & hellaswag\_vi \\
\midrule
\textbf{Llama-2-7B}          & 0.2588        & 0.2618        & 0.2634        & 0.3522        \\
\textbf{Llama-2-7B-Chat}     & 0.2594        & 0.2599        & 0.2627        & 0.3445        \\
\textbf{Llama-3-8B-Instruct} & 0.2759        & 0.2770        & 0.2791        & 0.4021        \\
\textbf{Vicuna-7B-v1.5}      & 0.2595        & 0.2644        & 0.2659        & 0.3556        \\
\textbf{Mistral-7B}          & 0.2633        & 0.2617        & 0.2733        & 0.3614        \\
\textbf{BayLing-1-7B}        & 0.2577        & 0.2576        & 0.2598        & 0.2831        \\
\textbf{BayLing-2-7B}        & 0.2585        & 0.2631        & 0.2645        & 0.3458        \\
\textbf{BayLing-2-8B}        & 0.2765        & 0.2807        & 0.2948        & 0.4025       \\
\bottomrule
\end{tabular}%
\end{table}

\begin{table}[h]
\centering\tiny
\caption{Numerical results of XNLI benchmark.}
\label{tab:model_xnl}
\begin{tabular}{lcccccc}\toprule
\textbf{Model}               & xnli\_bg & xnli\_de & xnli\_es & xnli\_sw & xnli\_ur & xnli\_zh \\
\midrule
\textbf{Llama-2-7B}          & 0.4249   & 0.4699   & 0.4064   & 0.3478   & 0.3357   & 0.3639   \\
\textbf{Llama-2-7B-Chat}     & 0.3723   & 0.4321   & 0.3956   & 0.3410   & 0.3410   & 0.3699   \\
\textbf{Llama-3-8B-Instruct} & 0.4518   & 0.4932   & 0.4863   & 0.3622   & 0.3454   & 0.4052   \\
\textbf{Vicuna-7B-v1.5}      & 0.4177   & 0.4663   & 0.4763   & 0.3446   & 0.3398   & 0.4056   \\
\textbf{Mistral-7B}          & 0.4309   & 0.4639   & 0.4406   & 0.3510   & 0.3414   & 0.3960   \\
\textbf{BayLing-1-7B}        & 0.4213   & 0.4474   & 0.4558   & 0.3450   & 0.3438   & 0.3614   \\
\textbf{BayLing-2-7B}        & 0.3594   & 0.4036   & 0.3936   & 0.3345   & 0.3422   & 0.3731   \\
\textbf{BayLing-2-8B}        & 0.4510   & 0.4972   & 0.4908   & 0.3655   & 0.3454   & 0.4096  \\
\bottomrule
\end{tabular}
\end{table}

\begin{table}[h]
\centering\tiny
\caption{Numerical results of multilingual ARC benchmark.}
\label{tab:model_arc}
\begin{tabular}{lccccccccc}\toprule
\textbf{Model}               & arc\_ar & arc\_eu & arc\_hi & arc\_hy & arc\_mr & arc\_ne & arc\_sr & arc\_uk & arc\_zh \\
\midrule
\textbf{Llama-2-7B}          & 0.0128  & 0.0079  & 0.0283  & 0.0036  & 0.0208  & 0.0214  & 0.0470  & 0.0565  & 0.0556  \\
\textbf{Llama-2-7B-Chat}     & 0.0034  & 0.0035  & 0.0128  & 0.0018  & 0.0078  & 0.0060  & 0.0180  & 0.0359  & 0.0214  \\
\textbf{Llama-3-8B-Instruct} & 0.1172  & 0.0360  & 0.0839  & 0.0245  & 0.0545  & 0.0496  & 0.1138  & 0.1437  & 0.1248  \\
\textbf{Vicuna-7B-v1.5}      & 0.0197  & 0.0053  & 0.0317  & 0.0036  & 0.0130  & 0.0111  & 0.0393  & 0.0582  & 0.0855  \\
\textbf{Mistral-7B}          & 0.0188  & 0.0097  & 0.0068  & 0.0009  & 0.0035  & 0.0043  & 0.0753  & 0.0804  & 0.0632  \\
\textbf{BayLing-1-7B}        & 0.0145  & 0.0062  & 0.0214  & 0.0018  & 0.0173  & 0.0137  & 0.0359  & 0.0325  & 0.0752  \\
\textbf{BayLing-2-7B}        & 0.0205  & 0.0053  & 0.0146  & 0.0027  & 0.0069  & 0.0120  & 0.0248  & 0.0214  & 0.0718  \\
\textbf{BayLing-2-8B}        & 0.1223  & 0.0518  & 0.0865  & 0.0255  & 0.0580  & 0.0667  & 0.1172  & 0.1600  & 0.1684 \\
\bottomrule
\end{tabular}
\end{table}

\end{document}